\documentclass[a4paperfleqn]{arXivCas-dc}

\usepackage{natbib}
\setcitestyle{authoryear,round,longnamesfirst}

\usepackage{pdflscape}
\usepackage{subcaption}
\usepackage{titlesec}
\usepackage{cleveref}
\usepackage{comment}
\usepackage{enumitem}
\usepackage{nicefrac}
\usepackage{hhline}
\usepackage{nicematrix}
\usepackage{tikz}
\usetikzlibrary{calc,intersections,shapes.symbols, shapes.geometric,positioning,arrows,arrows.meta,patterns,patterns.meta}

\newcommand{\operationalColor}{lime!80!black!20}
\newcommand{\processColor}{cyan!30}
\newcommand{\sceneryColor}{green!30}
\newcommand{\environmentColor}{yellow!30}
\newcommand{\objectsColor}{brown!30}
\newcommand{\propColor}{magenta!20}
\newcommand{\lodColor}{gray!60}

\makeatletter
\pgfdeclareshape{fieldoutline}
{
  \inheritsavedanchors[from=rectangle]
  \inheritanchor[from=rectangle]{center}
  \inheritanchor[from=rectangle]{north}
  \inheritanchor[from=rectangle]{south}
  \inheritanchor[from=rectangle]{west}
  \inheritanchor[from=rectangle]{east}
  \inheritanchor[from=rectangle]{north west}
  \inheritanchor[from=rectangle]{north east}
  \inheritanchor[from=rectangle]{south west}
  \inheritanchor[from=rectangle]{south east}

  \anchor{corner1}{\pgfpoint{0.1\wd\pgfnodeparttextbox}{0.15\ht\pgfnodeparttextbox}}
  \anchor{corner2}{\pgfpoint{0.5\wd\pgfnodeparttextbox}{0.15\ht\pgfnodeparttextbox}}
  \anchor{corner3}{\pgfpoint{0.96\wd\pgfnodeparttextbox}{0.05\ht\pgfnodeparttextbox}}
  \anchor{corner4}{\pgfpoint{0.95\wd\pgfnodeparttextbox}{0.25\ht\pgfnodeparttextbox}}
  \anchor{corner5}{\pgfpoint{0.85\wd\pgfnodeparttextbox}{0.975\ht\pgfnodeparttextbox}}
  \anchor{corner6}{\pgfpoint{0.6\wd\pgfnodeparttextbox}{0.8\ht\pgfnodeparttextbox}}
  \anchor{corner7}{\pgfpoint{0.15\wd\pgfnodeparttextbox}{0.98\ht\pgfnodeparttextbox}}
  \anchor{corner8}{\pgfpoint{0.1\wd\pgfnodeparttextbox}{0.75\ht\pgfnodeparttextbox}}
  \anchor{corner9}{\pgfpoint{0.15\wd\pgfnodeparttextbox}{0.5\ht\pgfnodeparttextbox}}

  \backgroundpath{
    \pgfpathmoveto{\pgfpoint{-0.1\wd\pgfnodeparttextbox}{0.15\ht\pgfnodeparttextbox}} 
    \pgfpathlineto{\pgfpoint{0.5\wd\pgfnodeparttextbox}{0.15\ht\pgfnodeparttextbox}}   
    \pgfpathlineto{\pgfpoint{0.96\wd\pgfnodeparttextbox}{0.05\ht\pgfnodeparttextbox}}   
    \pgfpathlineto{\pgfpoint{0.95\wd\pgfnodeparttextbox}{0.25\ht\pgfnodeparttextbox}}  
    \pgfpathlineto{\pgfpoint{0.85\wd\pgfnodeparttextbox}{0.975\ht\pgfnodeparttextbox}} 
    \pgfpathlineto{\pgfpoint{0.6\wd\pgfnodeparttextbox}{0.8\ht\pgfnodeparttextbox}}    
    \pgfpathlineto{\pgfpoint{0.15\wd\pgfnodeparttextbox}{0.98\ht\pgfnodeparttextbox}}  
    \pgfpathlineto{\pgfpoint{0.1\wd\pgfnodeparttextbox}{0.75\ht\pgfnodeparttextbox}}   
    \pgfpathlineto{\pgfpoint{0.15\wd\pgfnodeparttextbox}{0.5\ht\pgfnodeparttextbox}}   
    \pgfpathclose 
  }
}
\pgfdeclareshape{fieldoutlineStop}
{
  \inheritsavedanchors[from=rectangle]
  \inheritanchor[from=rectangle]{center}
  \inheritanchor[from=rectangle]{north}
  \inheritanchor[from=rectangle]{south}
  \inheritanchor[from=rectangle]{west}
  \inheritanchor[from=rectangle]{east}
  \inheritanchor[from=rectangle]{north west}
  \inheritanchor[from=rectangle]{north east}
  \inheritanchor[from=rectangle]{south west}
  \inheritanchor[from=rectangle]{south east}

  \anchor{corner1}{\pgfpoint{0.1\wd\pgfnodeparttextbox}{0.15\ht\pgfnodeparttextbox}}
  \anchor{corner2}{\pgfpoint{0.5\wd\pgfnodeparttextbox}{0.15\ht\pgfnodeparttextbox}}
  \anchor{corner3}{\pgfpoint{0.96\wd\pgfnodeparttextbox}{0.05\ht\pgfnodeparttextbox}}
  \anchor{corner4}{\pgfpoint{0.95\wd\pgfnodeparttextbox}{0.25\ht\pgfnodeparttextbox}}
  \anchor{corner5}{\pgfpoint{0.85\wd\pgfnodeparttextbox}{0.975\ht\pgfnodeparttextbox}}
  \anchor{corner6}{\pgfpoint{0.6\wd\pgfnodeparttextbox}{0.8\ht\pgfnodeparttextbox}}
  \anchor{corner7}{\pgfpoint{0.15\wd\pgfnodeparttextbox}{0.98\ht\pgfnodeparttextbox}}
  \anchor{corner8}{\pgfpoint{0.1\wd\pgfnodeparttextbox}{0.75\ht\pgfnodeparttextbox}}
  \anchor{corner9}{\pgfpoint{0.15\wd\pgfnodeparttextbox}{0.5\ht\pgfnodeparttextbox}}

  \backgroundpath{
    \pgfpathmoveto{\pgfpoint{-0.1\wd\pgfnodeparttextbox}{0.15\ht\pgfnodeparttextbox}} 
    \pgfpathlineto{\pgfpoint{0.5\wd\pgfnodeparttextbox}{0.15\ht\pgfnodeparttextbox}}   
    \pgfpathlineto{\pgfpoint{0.61\wd\pgfnodeparttextbox}{0.31\ht\pgfnodeparttextbox}}   
    \pgfpathlineto{\pgfpoint{0.46\wd\pgfnodeparttextbox}{0.35\ht\pgfnodeparttextbox}}  
    \pgfpathlineto{\pgfpoint{0.67\wd\pgfnodeparttextbox}{0.85\ht\pgfnodeparttextbox}} 
    \pgfpathlineto{\pgfpoint{0.6\wd\pgfnodeparttextbox}{0.8\ht\pgfnodeparttextbox}}    
    \pgfpathlineto{\pgfpoint{0.15\wd\pgfnodeparttextbox}{0.98\ht\pgfnodeparttextbox}}  
    \pgfpathlineto{\pgfpoint{0.1\wd\pgfnodeparttextbox}{0.75\ht\pgfnodeparttextbox}}   
    \pgfpathlineto{\pgfpoint{0.15\wd\pgfnodeparttextbox}{0.5\ht\pgfnodeparttextbox}}   
    \pgfpathclose 
  }
}

\def\tikz@auto@anchor{%
    \pgfmathtruncatemacro\angle{atan2(\pgf@y,\pgf@x)-90}
    \edef\tikz@anchor{\angle}%
}

\def\tikz@auto@anchor@prime{%
    \pgfmathtruncatemacro\angle{atan2(\pgf@y,\pgf@x)+90}
    \edef\tikz@anchor{\angle}%
}
\makeatother
\usepackage{siunitx}

\crefname{figure}{Fig.}{Figs.}
\crefname{table}{Table}{Tables}
\crefname{section}{Section}{Sections}
\crefname{enumi}{Scenario}{Scenarios}
\Crefname{figure}{Fig.}{Figs.}
\Crefname{table}{Table}{Tables}
\Crefname{section}{Section}{Sections}
\Crefname{enumi}{Scenario}{Scenarios}

\newcommand*\circled[1]{\tikz[baseline=(char.base)]{
            \node[shape=circle,draw,inner sep=0.7pt] (char) {#1};}}

\newcommand*\boxeded[2]{\tikz[baseline=(char.base)]{
            \node[shape=rectangle,draw,inner sep=1.2pt, fill=#2, rounded corners=0.1cm] (char) {~#1~\vphantom{Qp}};}}
            
\usepackage{multirow}
\usepackage[switch]{lineno}

\usepackage{amssymb}
\usepackage{pgfplotstable}
\pgfplotsset{compat=1.16}
\usetikzlibrary{patterns}

\renewcommand\normalsize{\fontsize{10pt}{10pt}\selectfont}

\titleformat*{\section}{\normalsize\bfseries}
\titleformat*{\subsection}{\normalsize\bfseries}
\titleformat*{\subsubsection}{\normalsize\bfseries}
\titleformat*{\paragraph}{\normalsize\bfseries}
\titleformat*{\subparagraph}{\normalsize\bfseries}

\begin{document}
\sloppy
\let\WriteBookmarks\relax
\def\floatpagepagefraction{1}
\def\textpagefraction{.001}

\hypersetup{
  linkcolor  = [RGB]{33,150,209},
  citecolor  = [RGB]{33,150,209},
  urlcolor   = [RGB]{33,150,209},
}

\title[mode = title]{Toward an Agricultural Operational Design Domain: A Framework}
\title[mode=sub]{Enabling Farm Autonomy by Logical Scenarios and Ag-ODD Derivation}

\shorttitle{Toward an Agricultural Operational Design Domain: A Framework}
 \shortauthors{M. Felske et al.}
  
  \author[1]{Mirco Felske}[type=editor,
    orcid=0009-0004-1740-5562]
  \ead[1]{mirco.felske@claas.com}
  
\author[1]{Jannik Redenius}[type=editor,
    orcid=0000-0002-6964-0968]
  \ead[2]{jannik.redenius@claas.com}
  \address[1]{CLAAS E-Systems GmbH, Dissen a.T.W., Germany}

  \author[2]{Georg Happich}[type=editor,
    orcid=0009-0001-4975-3703]
  \ead[2]{georg.happich@hs-kempten.de}
  \address[2]{Department of Mechanical Engineering, Kempten University of Applied Sciences, Kempten, Germany}

\author[3]{Julius Schöning}[type=editor,
    orcid=0000-0003-4921-5179]
  \ead[3]{j.schoening@hs-osnabrueck.de}
  \cormark[1]
  \address[3]{Faculty of Engineering and Computer Science, Osnabrück University of Applied Sciences, Osnabrück, Germany}

 \begin{abstract}[S U M M A R Y]
The agricultural sector increasingly relies on autonomous systems that operate in complex and variable environments. Unlike on-road applications, agricultural automation integrates driving and working processes, each of which imposes distinct operational constraints. Handling this complexity and ensuring consistency throughout the development and validation processes requires a structured, transparent, and verified description of the environment. However, existing Operational Design Domain (ODD) concepts do not yet address the unique challenges of agricultural applications.

Therefore, this work introduces the Agricultural ODD (Ag-ODD) Framework, which can be used to describe and verify the operational boundaries of autonomous agricultural systems. The \textbf{Ag-ODD Framework} consists of three core elements.
First, the \textbf{Ag-ODD} description concept, which provides a structured method for unambiguously defining environmental and operational parameters using concepts from ASAM Open ODD and CityGML.
Second, the  \textbf{7-Layer Model} derived from the PEGASUS 6-Layer Model, has been extended to include a process layer to capture dynamic agricultural operations.
Third, the \textbf{iterative verification process} verifies the Ag-ODD against its corresponding logical scenarios, derived from the 7-Layer Model, to ensure the Ag-ODD's completeness and consistency.

Together, these elements provide a consistent approach for creating unambiguous and verifiable Ag-ODD. Demonstrative use cases show how the Ag-ODD Framework can support the standardization and scalability of environmental  descriptions for autonomous agricultural systems.
 \end{abstract}
  \maketitle

\section{Introduction}
In recent years, technological progress has profoundly transformed the potential of automation in off-road domains such as agriculture. Advances in machine perception using neural networks, high-performance computing on edge devices, and networked machinery have enabled increasingly autonomous operation in complex, unstructured environments \citep{Hindman2024}. Automation that has been limited to controlled industrial settings is now moving into open fields and dynamic agricultural contexts.

\begin{figure*}
    \centering
    \input{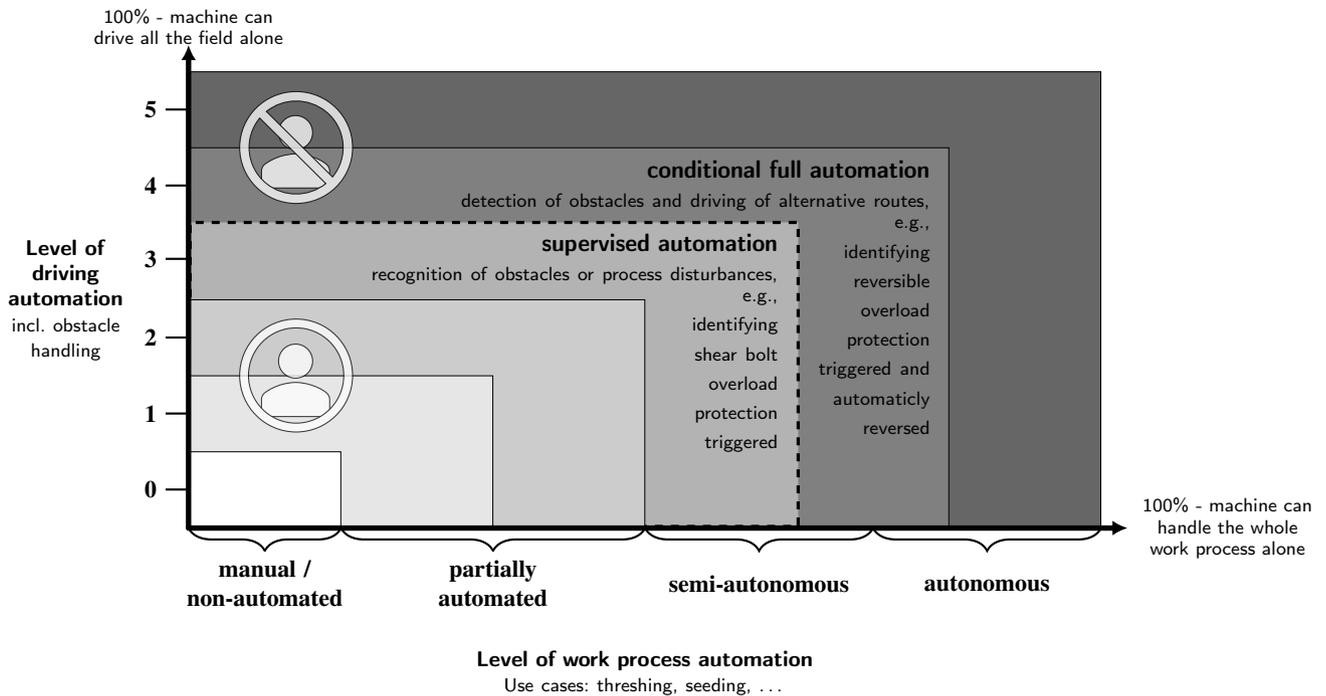}
    \caption{Multiple dimensions of automation in agriculture: The individual degree of automation can be defined both for the driving task according to \citet{SAEJ30162021} and for the working process according to \citet{ISO18497} of the machine or device. 
    }    
   \label{fig:Automation_of_driving+working}
\end{figure*}

Compared to other domains, e.g., the automotive sector, agricultural automation encompasses more than the driving task alone. While there exists a clear interpretation of automotive's six automation levels \citep{SAEJ30162021}, assessing that in an agricultural context requires a definition of automation in both functions and modes \citep{ISO18497}. Mapping these approaches has been performed linear in \citet{Streitberger2018} and multidimensional in \citet{Schoening2025}. 
As a merge of these approaches driving automation and the automation of working processes can be scaled in a two-dimensional planar space as shown in \cref{fig:Automation_of_driving+working}. This allows for an easy ranking of different machines and machine types: The more the specific system is allocated to the upper right, the higher the overall automation level is. Note that \citet{ISO18497} stated that they deliberately did not choose the \citet{SAEJ30162021} levels of automation as a basis. 

The vehicle's motion is typically subordinate to the agricultural process itself, whereby each is imposing specific requirements on speed, trajectory, and timing. Moreover, agricultural environments vary widely: soil conditions, crop types and growth stages, and the presence of natural obstacles make it difficult to consistently describe the range of situations in which autonomous systems must operate.

Following these constraints, several recent publications have highlighted the urgent need for a clearly defined Operational Design Domain (ODD) concept tailored to agricultural applications \citep{Happich2025,Schoening2024,Happich2025a}. Such an Agricultural ODD (Ag-ODD) would provide a structured way to specify environmental and operational boundaries for autonomous agricultural systems - whether for interpreting performance limitations \citep{Baillie2020}, constraining validation efforts \citep{Komesker2024}, ensuring legal conformity \citep{Kruse2024}, and supporting simulation-based testing \citep{Tauber2024}. Despite growing attention to this topic, there is still no standardized framework or shared, harmonized understanding of how to define, represent, and apply such information practice.

Therefore a structured approach is required to manage this complexity. Validation of autonomous functions across realistic environmental variations, simulating consistent operating conditions and trace parameter dependencies throughout the development process will become increasingly difficult with absence of such a framework. These challenges underscore the need for a transparent, unified structure linking environmental description, system design, and verification logic. 

To address this need, this paper proposes a framework to specify an Ag-ODD. This framework provides a unique definition of environmental descriptions and meets the complex requirements of agricultural processes. To this end, the paper introduces the four components of the Ag-ODD Framework, which are used together. The first component is the use case, which is required to derive the second and third components: the Ag-ODD and the 7-Layer Model which is in turn based on the 6-Layer Model from PEGASUS. The fourth component is the iterative verification process, which verifies the Ag-ODD and the logical scenarios derived by the 7-Layer Model. The Ag-ODD Framework, the core focus of this work, is presented in \cref{sec:Ag-ODD_definition} and demonstrated using constructed examples in \cref{sec:examples}.
Prior to this, \cref{sec:SOTA} analyses existing approaches, and \cref{sec:synopsis and synthesis} uses this information to identify the missing attributes and properties required for an Ag-ODD to be structured and unambiguous.

\section{Related Standards and Frameworks}\label{sec:SOTA}
The development and safe deployment of automated systems is contingent upon a foundation of established standards and frameworks. This section explores the key methodologies employed to define and validate the operational boundaries of these systems, focusing on the critical concept of the Operational Design Domain (ODD) and its implementation across various applications.

\subsection{Operational Design Domain --- On-Road}
The Operational Design Domain (ODD) is central to understanding the capabilities and limitations of automated driving systems on public roads. This section delves into the core principles of the ODD, as defined by \citet{SAEJ30162021} and further refined by methodologies like PEGASUS and ASAM OpenODD, examining how these frameworks ensure safe and reliable operation within specified conditions.

\subsubsection{SAE~J3016}
Revised in 2021, the \citet{SAEJ30162021} standard serves as a pivotal framework for driving automation on public roads. It provides a taxonomy of six levels, ranging from no automation (Level 0) to full automation (Level 5). The ODD is a fundamental component of this taxonomy, delineating the parameters within which a driving automation system operates. The ODD encompasses environmental, geographical, and temporal limitations, as well as the presence or absence of specific traffic or roadway characteristics \citep{SAEJ30162021}.

The ODD is particularly significant for Levels 3 to 5, when the Automated Driving System (ADS) carries out the entire Dynamic Driving Task (DDT). The ODD helps developers and users understand the system's limitations and capabilities. The ODD ensures the ADS operates within its designated parameters. For example, a Level 4 ADS may be designed to operate within a specific geographic area or under certain weather conditions. In contrast, a Level 5 ADS is designed to operate under all driver-manageable on-road conditions.

The ODD is integral to defining the Dynamic Driving Task (DDT).
The DDT encompasses all real-time operational and tactical functions required to operate a vehicle in on-road traffic. The DDT includes subtasks such as controlling the vehicle's lateral and longitudinal motion, monitoring the driving environment, and detecting and responding to objects and events. The ODD ensures these subtasks are executed within the system's predetermined operational limits.

Additionally, the ODD plays a pivotal role in the DDT fallback response, which must be implemented when the system cannot continue executing the DDT due to a failure or termination. At Levels 3 to 5, the system must demonstrate the ability to autonomously execute the DDT fallback and attain a minimal risk condition. The ODD determines the timing and method of the fallback process, ensuring the system can transition to a minimal risk condition seamlessly if necessary.

The \citet{SAEJ30162021} standard emphasizes the importance of clear definitions and roles, which aids in developing laws, policies, and technical specifications. The ODD concept promotes international collaboration by adhering to ISO standards, ensuring uniform categorization of driving automation systems. The standard's influence is evident through its adoption by organizations such as the OICA, which promotes a unified approach to Automated Driving (AD) technologies.

The ODD is a foundational component of the \citet{SAEJ30162021} standard. It serves as a framework for understanding the operational limits and capabilities of driving automation systems. It provides the industry with guidance as it advances AD technologies while maintaining clarity and consistency. This ensures that systems operate safely and effectively within their designed parameters.

\subsubsection{PEGASUS Method}\label{sec:pegasus}
The PEGASUS method is a systematic approach designed to standardize the development and evaluation of Advanced Driver Assistance System (ADAS) and ADS functions, thereby ensuring consistency and independence from specific tools, manufacturers, or implementation details \citep{PPO2021}. This approach is intended to ensure the comparability and reproducibility of results within the industry. The method is divided into five steps.

The first step involves data processing, where logical scenarios are identified from abstract use-case knowledge and real-world traffic data. These scenarios, which are frequently textual in nature, are consolidated into a uniform format for subsequent database utilization.

The second step involves defining requirements by deriving functional and behavioral expectations from the scenarios, along with evaluation criteria for system performance. These criteria are aligned with safety frameworks such as \citet{ISO26262} and \citet{ISO21448} Safety Of The Intended Functionality (SOTIF).

The third step involves storing the scenarios and evaluation criteria in a structured database using the PEGASUS 6-Layer Model. This model organizes information into six categories: road topology, infrastructure, temporary changes, dynamic objects, environmental conditions, and digital information. The 6-Layer Model standardizes not only scenario descriptions but also the definition of the Operational Design Domain, fostering a shared understanding among manufacturers, regulators, and test organizations. The database maps scenarios into a multidimensional parameter space, utilizing metrics and statistics to facilitate systematic coverage.

The fourth step involves assessing the ADAS and AD function by first executing the scenarios in simulation, then on proving grounds, and finally in real traffic. This assessment ensures traceability across test stages.

The final step involves safety argumentation, which compares the collected test evidence to predefined safety claims.

\subsubsection{Association for Standardisation of Automation Open Operational Design Domain}\label{sec:OpenODD}
In Association for Standardisation of Automation (ASAM) OpenODD \citep{ASAM2021}, the term ODD refers to the set of operating conditions under which an ADAS and ADS is designed to operate safely and reliably. The aforementioned conditions encompass a multitude of factors, including but not limited to environmental elements, infrastructural characteristics, geographical attributes, and dynamic components such as other road users. These definitions align with the established criteria outlined in \citet{SAEJ30162021} and \citet{ISO34503}.

The standard uses a hierarchical, ontology-based model via OpenXOntology and the OpenSCENARIO DSL, making ODD both human- and machine-readable. Within this structure, an ODD is organized into three top-level categories: Scenery, which includes road types and infrastructure; Environment, which encompasses aspects such as weather and lighting; and Dynamic Objects, which represent elements like vehicles and pedestrians. Each category can be expanded with multiple sub-attributes in a layered hierarchy.

ASAM OpenODD introduces two definition modes: restrictive, where anything not specified is excluded, and permissive, where anything not specified is included. These modes can be applied either globally or to individual attributes, providing precise control. This ambiguity highlights the importance of clear boundaries to prevent ``fuzzy'' boundaries in the ODD. Defined boundaries enable precise measurement and a reliable assessment of test coverage. The standard requires manufacturers to supply a defined ODD, and it holds operators responsible for ensuring that the ADS and ADAS are only used within the defined ODD.

\subsection{Operational Design Domain --- Off-Road}
Extending beyond paved roads, defining the ODD for off-road applications presents unique challenges due to dynamic and often unpredictable environments. This section examines how standards from organizations like EMESRT and NATO address these complexities, providing frameworks for specifying operational limits in demanding terrains and conditions.

\subsubsection{Earth Moving Equipment Safety Round Table Performance Requirement 5A}
The Earth Moving Equipment Safety Round Table (EMESRT) Performance Requirement 5A (PR-5A) \citep{EMESRTcircledR2024} provides a methodical approach to specifying the ODD for vehicle interaction systems in the mining industry. The PR-5A is based on EMESRT Design Philosophy 5, which focuses on machine operation and control. This philosophy delineates the expected operational contexts under which collision avoidance systems should perform reliably. These contexts encompass environmental conditions, such as dust, lighting, terrain, and vehicle states, including reversing, startup, shutdown, as well as complex dynamic interactions with pedestrians, other vehicles, and static infrastructure. The PR-5A system facilitates a functional interpretation of ODD boundaries by identifying specific Potential Unwanted Events (PUEs), including vehicle-to-person or vehicle-to-equipment interactions. This specification enables system developers to align detection, alerting, and intervention capabilities with credible, real-world operational constraints inherent in both surface and underground mining environments.

The PR-5A framework is based on the EMESRT 9-Layer Model of Control Effectiveness, which outlines safety functions across a hierarchy, ranging from foundational operational practices to advanced automation. The upper three layers---situational awareness (Level 7), advisory controls (Level 8), and intervention controls (Level 9 )---are where technology-based systems must be designed to operate within a defined ODD. These levels are indicative of increasing system autonomy and escalating requirements for sensing accuracy, decision-making latency, and contextual awareness. A variety of factors, including vehicle speed, closure rate, sensor field of view, operator workload, and proximity to other actors or infrastructure, influence the ODD parameters at each level. The PR-5A highlights the site-dependency and dynamism of these parameters, thereby underscoring the need for flexible and validated ODD modeling. Such modeling must account for the technological limitations and human factors influencing system performance in hazardous work zones.

The incorporation of a series of Functional Performance Scenario Storyboards (FPSS) within PR-5A is instrumental in facilitating the application and validation of ODD specifications. These scenarios offer detailed visualizations of operational situations that systems must interpret and respond to, including scenarios such as tailgating, blind approach, pedestrian ingress, and operation in congested zones. Each storyboard outlines the anticipated system behavior, identifies pertinent failure modes, and specifies environmental variables that may impact system performance. The incorporation of these scenarios into the design process enables PR-5A to facilitate empirical validation of ODD boundaries through use-case testing and site-specific configuration. This scenario-based methodology strengthens the connection between theoretical system capabilities and their practical deployment in safety-critical environments. Consequently, PR-5A provides a replicable model for ODD definition and validation, thereby facilitating the development of safety technologies and advancing research into autonomous and semi-autonomous vehicle interaction systems.

\subsubsection{AMSP-06 for Ground Vehicle Mobility}\label{sec:AMSP}
The NATO Standard \citet{NSO2021}, which delineates the architectural specifications for the Next-Generation NATO Reference Mobility Model (NG-NRMM), implicitly addresses the ODD by integrating detailed environmental, vehicle, and terrain factors into its modeling framework. In contrast to conventional vehicle mobility assessments, which are predicated on deterministic assumptions, \citet{NSO2021} employs a probabilistic approach that acknowledges the variability and complexity inherent in operational environments. The NG-NRMM integrates geospatial data, 3D vehicle dynamics, terramechanics, and autonomous control simulations to construct a multidimensional ODD. This ODD accounts for terrain type, soil strength, elevation, weather, and mobility-constraining obstacles. These attributes enable NATO member states to evaluate not only the capacity of a vehicle to traverse a designated terrain, but also the probability of successful traversal under specified conditions, including scenarios where sensing or control performance is degraded, or the environment is uncertain or evolving.

The primary contribution of \citet{NSO2021} to the definition of ODD is rooted in its layered modularity and explicit coupling of vehicle models with terrain and environmental data. The key variables that define the ODD include terrain composition, such as soil classification, roughness, and moisture content; topographical features, including slope and elevation; and contextual overlays, such as vegetation, snow, or urban debris. The simulation of vehicle systems is achieved through the integration of multi-body dynamics with terramechanics, thereby reflecting realistic interactions between running gear and deformable terrain. These interactions delineate performance boundaries, including speed made good, trafficability in GO and NOGO conditions, and motive efficiency. Each parameter is mapped across spatial grids derived from high-resolution GIS data. This formalization enables scenario-specific assessments, wherein the operational limits of vehicles are evaluated in context---an essential capability for planning autonomous or semi-autonomous operations in NATO theaters.

The incorporation of uncertainty quantification and scenario-specific overlays into \citet{NSO2021} endows it with a flexible structure for stress-testing ODD boundaries across terrain and mission profiles. Terrain inputs, for instance, may be designated as measured, inferred, or notional, with associated uncertainty metadata encoded for each attribute. The integration of elevation data, soil strength models, and dynamic obstacles can be configured to simulate nominal and edge-case scenarios. This approach not only facilitates system validation and verification (V\&V) but also informs real-time operational planning tools by generating probabilistic mobility maps. In this sense, \citet{NSO2021} operationalizes the concept of an ODD not as a fixed envelope, but as a context-dependent, quantifiable domain whose limits are discoverable and adaptable through simulation. This positions \citet{NSO2021} as a foundational reference for research into adaptive vehicle behavior, mission planning under uncertainty, and autonomous navigation in complex, contested environments.

\subsection{Markup Languages of Operational Design Domain }
Formalizing the ODD requires robust methods for representing and exchanging information about operational environments. This section explores various markup languages---including CityGML, OpenPLX, ASAM OSI, and ISTVS standards---and their potential for encoding and utilizing ODD data in simulation, testing, and real-world deployment of automated systems.

\subsubsection{City Geography Markup Language}\label{sec:CityGML}
City Geography Markup Language (CityGML, \citep{CityGML2025}, a semantic 3D city model standard developed by the Open Geospatial Consortium (OGC), facilitates the structured representation of urban environments, encompassing their geometry, semantics, topology, and appearance. Initially published as version 1.0 in 2008 and subsequently revised in version 2.0 \citep{Groger2008}, CityGML has evolved into a robust framework for spatial data modeling that extends well beyond mere visualization.

The most recent release, CityGML 3.0, introduces a clear separation between the conceptual model and its encodings, like eXtensible Markup Language (XML), GML, JavaScript Object Notation (JSON), and Resource Description Framework (RDF), reflecting a shift toward flexible and interoperable data exchange \citep{OGC_Document_20-010}. The system is characterized by a modular structure, which is organized around thematic modules such as Building, Relief, Land Use, Vegetation, and Transportation. The extensibility of these modules is achieved through the implementation of Application Domain Extensions (ADEs), a feature that facilitates the adaptation of CityGML to meet the specific requirements of a given domain while maintaining semantic and topological consistency \citep{Kolbe2009}.

In the context of defining ODD for autonomous agricultural systems, CityGML offers significant potential. It is essential to note that the LandUse, Vegetation, and Relief modules offer semantically rich constructs that can be directly mapped to terrain classifications, crop zones, and biomass parameters, respectively. Recent advancements, such as the Vegetation ADE \citep{PetrovaAntonova2024} and the dynamizer module \citep{Kutzner2020}, facilitate the temporal modeling of plant growth, environmental factors, and machine operations---parameters that are pivotal to numerous agricultural ODD (Ag-ODD) definitions. Moreover, the Level-of-Detail (LoD) concept \citep{Biljecki2016} provides a context-sensitive abstraction in modeling. When transferred to the domain of agricultural use, this concept demonstrates significant potential. For instance, simulation, with its emphasis on specific aspects, manifests at varied levels of abstraction.

In summary, CityGML, supported by its extensible semantic core and well-established encoding logic, provides a transferable framework for modeling and parameterizing Ag-ODD. It facilitates the formal representation of environmental and infrastructural conditions, while concurrently providing support for rule-based constraints through ADE-based semantic enhancements. The LoD concept enables context-sensitive modeling. This concept renders CityGML a compelling candidate for incorporation into comprehensive modeling ecosystems that aspire to facilitate the safe and context-aware deployment of autonomous agricultural machinery. However, CityGML has been designed for a much broader range of applications than those related to the parametrization of agricultural use cases. It has been determined that the aforementioned methodology does not appear to satisfy the requirement for a lean and fast descriptive methodology. However, it has been observed that several concepts, including LoD, are pertinent for utilization in an Ag-ODD data model.

\subsubsection{OpenPLX}
OpenPLX \citep{ASA2025} is a domain-extensible modeling language introduced in 2024 by Algoryx Simulation AB, a Swedish company specializing in real-time physics simulation for mechanical systems through its platform, AGX Dynamics \citep{Algoryx2024a}. The development of this technology was initiated as part of an open-source initiative to establish a connection between high-fidelity physics simulation, modular autonomy logic, artificial intelligence components, and structured system modeling. The language is positioned as a declarative alternative to traditional simulation configuration formats, enabling formal system representations that are both simulation-native and control-aware.

OpenPLX is predicated on a component-based trait system that facilitates the definition of reusable, parameterized model elements. Traits are defined as the characteristics that define an entity's physical properties, logical constraints, signal behavior, and operational parameters. These are composed into object hierarchies using simple, declarative syntax constructs, such as ``is'', ``becomes'', etc., allowing for hierarchical modeling of complex systems \citep{OpenPLX_Docs}. One of the system's key features is its direct integration with CAD and URDF models, allowing for live-synchronized geometric references without the need for conversion or duplication \citep{ASA2025}. This renders OpenPLX particularly well-suited to workflows that integrate mechanical design with simulation and autonomy testing.

In comparison to established domain-specific formats, such as those delineated by ASAM, see \cref{sec:ASAM}. OpenPLX occupies a distinct niche within the industry. While the ASAM standards emphasize interoperability and formal validation artifacts for automotive domains \citep{ASA2025}, OpenPLX focuses on system-level simulation composition, prioritizing extensibility across domains and the runtime integration of signal interfaces, event triggers, and control logic. Its orientation towards reuse and modularity renders it well-suited for modeling heterogeneous systems beyond the road traffic context, including robotics, industrial automation, and mobile machinery.

In the context of agricultural automation, OpenPLX presents initial potential as a framework for simulation and testing within Ag-ODD. The aforementioned ODD are typically defined by parameters such as soil type, slope, weather, lighting, GNSS availability, and machine capability. These parameters can be encoded as reusable bundles of traits. To illustrate, a seeding operation may be constrained by factors such as loamy soil, daylight hours, the presence of RTK-GNSS, and a maximum slope of 5 degrees. These parameters can be formalized in OpenPLX using the same trait-based structure employed for robotics or drive train modeling. In any case, in comparison with the other frameworks that have been discussed in this work, OpenPLX offers a limited methodology for defining these ODD.

OpenPLX's signal interface and runtime logic enable direct validation of ODD constraints during both simulation and operation. Environmental or system violations, such as terrain slope exceeding defined thresholds, have the potential to trigger model-level responses. This capability is particularly valuable for virtual validation, certification preparation, or anomaly detection. The capacity of the language to integrate environmental modeling, machine capability, and behavioral rules into a unified executable model provides a pragmatic approach to achieving traceable and machine-readable Ag-ODD specifications. These specifications are a prerequisite for the broader deployment and regulation of autonomous agricultural machines.

\subsubsection{ASAM Open Simulation Interface}\label{sec:ASAM}
The ASAM Open Simulation Interface (OSI) is a standardized interface that describes environments, scenarios, and the exchange of sensor data between simulation components \citep{ASAM2024}. The OSI model delineates top-level messages, including GroundTruth, SensorView, and SensorData, that facilitate consistent information exchange across disparate simulation frameworks that support OSI. This consistent information exchange enables the reusability of scenarios irrespective of the underlying tools.

In the OSI model, environments are delineated through the OSI environment message and associated structures. These structures incorporate elements such as weather conditions, including precipitation, fog, and snow, lighting, shadows, occlusions, and other physical effects that influence sensor perception. The OSI sensor model is capable of producing low-level outputs, including point clouds and images, as well as high-level perception data, such as classified objects, detected features, and tracks. It has been demonstrated that high-level models possess the capacity to simulate filtering, feature extraction, and the fusion of multiple sensors, thereby reducing uncertainty. The logical model integrates multiple sensor perspectives to generate a unified dataset, thereby enhancing perception.

The traffic participant model delineates dynamic entities whose state is subject to change, encompassing vehicles, pedestrians, cyclists, and other moving objects. The text further elaborates on the subjects' positions, motions, and attributes. The structuring of data in this manner enables OSI to facilitate modular, reproducible, and framework-independent simulations for testing and validating automated driving systems.

\subsubsection{International Society for Terrain Vehicle System Terrain Vehicle Standards}
The 2020 standards established by the International Society for Terrain Vehicle Systems (ISTVS) function as a foundational reference for defining and normalizing key terrain vehicle interaction variables \citep{ISTVS2020}. While these standards do not explicitly use the term ODD, they provide critical building blocks for its construction by cataloging standardized terminology, measurement procedures, and metrics that span soil properties, vehicle components, mobility performance, and terramechanics testing devices. For researchers seeking to formalize an ODD for autonomous or semi-autonomous ground vehicles, the ISTVS standards offer the vocabulary and experimental protocols necessary to characterize the environmental and vehicular boundary conditions under which a mobility system is expected to function reliably.

The ODD in terrain–vehicle systems is predicated on the standardized terrain and mobility metrics defined in the ISTVS guidance. The standards encompass precise definitions of soil parameters, such as moisture content, bulk density, cohesion, friction angle, and vehicle attributes, including tire dimensions, slip ratios, and performance indicators, including traction force, sinkage, and drawbar pull. These parameters can be used to define the spatial and environmental limits of the ODD. For example, they can be used to specify soil types with acceptable cohesion ranges, maximum slippage thresholds, or sinkage depth limits. Through the integration of glossary definitions with standard test procedures, researchers can derive formal ODD constraints, including acceptable terrain roughness and moisture conditions, permissible wheel slip levels, sensor visibility thresholds, and mobility performance envelopes.

Furthermore, the ISTVS standards facilitate structured, scenario-based ODD validation through standardized test methods and equipment recommendations. A thorough exposition of soil-vehicle interaction trials is furnished herein, encompassing sinkage tests, shear ring tests, and mobility evaluation---for instance, slip and traction measurements. These trials are described with procedural rigor, ensuring consistency in their replication across research facilities. Consequently, the ODD is rendered a quantifiable domain, defined not solely by nominal parameter values but also validated through empirical data under defined test conditions. The alignment of controlled testing with definitional standardization supports a defensible and replicable ODD specification methodology for academic research and industry design alike.

\section{Synopsis and Synthesis for Agricultural Applications}\label{sec:synopsis and synthesis}

\Cref{sec:SOTA} analyzed a variety of promising candidates for the set-up, definition, modeling, and interpretation of an ODD in an agricultural context. It is essential to note that specific individual approaches can be implemented directly in an agricultural context, while others require adaptation for practical application. For instance, the 6-Layer Model is an excellent tool for describing logical ODD scenarios without omitting information. 

\subsection{Meta-Analysis's Résumé}\label{sec:meta_analysis_resumee}

Although the work is not intended to be a meta-analysis on existing ODD standards and definition methods, a brief summary will help understanding the work's course. In summary, none of the standards or frameworks presented in \cref{sec:SOTA} comprehensively meet the diverse requirements of the agricultural machinery and equipment industry for defining an Ag-ODD. Anyhow, due to its wide acceptance in the automotive industry, the SAE 3016's ODD definition is a promising candidate for adaption towards agricultural use cases. Beside that, the analysis emerged several methods and analogies which are advantageous extensions.

In a manner analogous to CityGML, ASAM OpenODD is predicated on a hierarchical structure that has the potential to be advantageous for agricultural ODD, as its LoD concept allows both a context and user-centric environmental description. In addition, the restrictive and permissive modes imply a capacity for clear delineation of ODD's boundaries. Although not stated as a mandatory requirement, a framework under development is expected to provide interfaces for OpenOSI. This would considerably enhance the framework's usability and facilitate its expeditious integration into existing simulation environments. With the goal of adopting existing frameworks towards an Ag-ODD, these promising artifacts form the relevant elements of choice; all extracted and based on standards driven by the automotive industry.

\subsection{Synopsis}\label{sec:synopsis}

The primary function of an ODD is to delineate the operational environment within which a function is expected to perform reliably. Consequently, an ODD must be specified in a manner that facilitates verification and, ideally, automated simulation. The candidates summarized in \cref{sec:meta_analysis_resumee} primarily originate in the automotive sector. However, the agricultural sector differs from the automotive sector in three fundamental respects:

\begin{itemize}
    \item The \textbf{geometry and dimensions} of agricultural equipment and machinery differ from those in the automotive industry. This includes that both the shape of, e.g., self-propelled vehicles such as harvesters and tractors may differ, as well as the width of, e.g., a pulled-type sprayer, which may exceed \SI{48}{\meter}.
    \item  Agricultural operations invariably entail a \textbf{dynamic process} that engenders alterations in the field; e.g., cultivation, mowing, and chopping are changing the state of the field continuously. Consequently, Ag-ODD specifications mandate precise descriptions of processes within the operational area itself rather than across long distances. 
    \item  Agricultural machinery generally operate at \textbf{speeds} of typically $\leq$~\SI{20}{\kilo\meter\per\hour}, enabling it to come to a complete halt within a short distance.
\end{itemize}

While in an automotive perspective the driving task is the most relevant focus for modeling, in an agricultural context the driving is just an instrumental value, an adequate means for a broader subject. For a variety of agricultural use cases, the working task, e.g., a rye harvesting process, its unique understanding, and the relevant modeling are of highest interest. Further addressing of this will happen later in \cref{sec:synthesis} in the form of bullet points.

\subsection{Synthesis}\label{sec:synthesis}

\begin{figure*}[H]
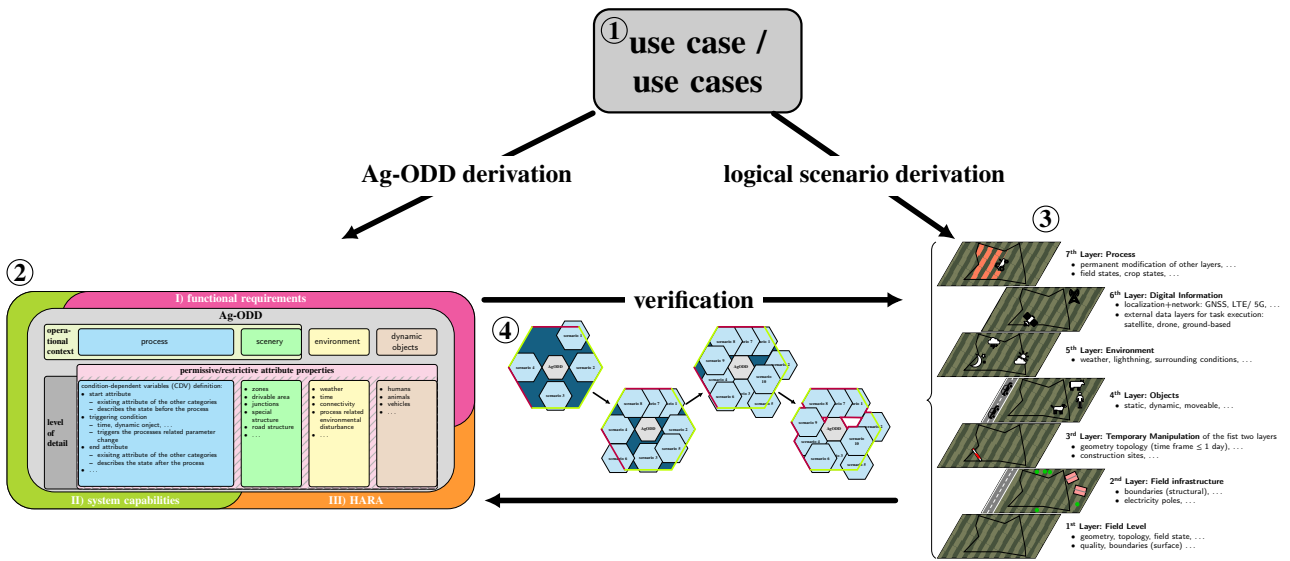

    \centering
  \resizebox{\linewidth}{!}{
 \begin{tikzpicture}

    \node[] (Ag-ODD){
        \resizebox{!}{2.75cm}{
        \begin{tikzpicture}
                 \input{Ag-ODD}
        \end{tikzpicture}}
        };
    \node[right = 0.15cm of Ag-ODD] (veri){
        \resizebox{!}{2.0cm}{
        \begin{tikzpicture}[
            hexagon/.style={regular polygon, regular polygon sides=6, draw=none, fill={rgb,255:red,21; green,95; blue,130}, inner sep=0cm,text width=2.4cm, minimum size=2.4cm, font=\scriptsize, align=center},
            hexagonODD/.style={regular polygon,regular polygon sides=6,draw, fill={rgb,255:red,224; green,224; blue,224},inner sep=0cm, text width=0.7cm, minimum size=0.7cm, align=center, font={\bf \tiny}},
            scenario/.style={regular polygon,regular polygon sides=6,draw, fill={rgb,255:red,193; green,228; blue,245},inner sep=0cm, text width=0.85cm, minimum size=0.85cm, align=center, font={\bf \tiny}},
            rl/.style={draw=purple,line width=2pt, line cap=round},
            gl/.style={draw=lime,line width=2pt, line cap=round},
            optl/.style={draw,line width=2pt, line cap=round, -{Latex[length=2mm, width=2mm]},shorten <=0.2cm, , shorten >=0.2cm},
            ]
        \input{iterative_ODD_verification}
        \end{tikzpicture}}
        };
    \node[right = 0.15cm of veri] (pega){
        \resizebox{!}{4.0cm}{
        \begin{tikzpicture}[
            field/.style={trapezium, trapezium left angle=60, trapezium right angle=120, draw, fill={rgb,255:red,71; green,82; blue,60}, text width=2.5cm, text height=1.5cm,inner sep=0cm, postaction={pattern={Lines[angle=60,distance=1.2mm,  line width=0.6mm]},pattern color={rgb,255:red,119; green,124; blue,90}},},
            fieldT/.style={text width=6.cm, font=\footnotesize,},
            fieldO/.style={draw, thick, fieldoutline, text width=2.0cm, text height=1.5cm,inner sep=0cm,},
            fieldS/.style={trapezium, trapezium left angle=60, trapezium right angle=120, text width=0mm, text height=1.5cm, inner xsep=-2.5mm, inner ysep=0mm,fill=gray},
            ]
         \input{ag-PEGASUS}
         \end{tikzpicture}
        }
        };
        \node(uc)[above = 2.5cm of veri, draw, thick, fill=black!20,text width=2.4cm, minimum height=4em, align=center, rounded corners=0.2cm, font={\bf \large}] {use case / use cases};
            
       \draw[draw,line width=2pt, line cap=round, -{Latex[length=2mm, width=2mm]}] (uc)-- node[midway, fill=white, font={\bf}]{Ag-ODD derivation}($(uc)+(-4.5,-2.25)$);
       \draw[draw,line width=2pt, line cap=round, -{Latex[length=2mm, width=2mm]}] (uc)-- node[midway, fill=white, font={\bf}]{logical scenario derivation}($(uc)+(3.25,-2.25)$);
       \draw[draw,line width=2pt, line cap=round, -{Latex[length=2mm, width=2mm]}] ($(veri.north west)+(-0.15,0.15)$)-- node[midway, fill=white, font={\bf}]{verification}($(veri.north east)+(0.05,0.15)$);
       \draw[draw,line width=2pt, line cap=round, -{Latex[length=2mm, width=2mm]}] ($(veri.south east)+(-0.05,-0.15)$)--($(veri.south west)+(-0.15,-0.15)$);
       
       \node[draw, below right = 0.4cm and 0.15cm of uc.north west, anchor=south west, inner sep=0cm, circle, minimum height=1em, font=\bf] (no1){1};
       \node[draw, below right = -0.0cm and 0.2cm of Ag-ODD.north west, anchor=south west, inner sep=0cm, circle, minimum height=1em, font=\bf] (no2){2};
       \node[draw, below left = -0.0cm and 0.75cm of pega.north, anchor=south, inner sep=0cm, circle, minimum height=1em, font=\bf] (no3){3};
       \node[draw, below right = 0.35cm and 0.0cm of veri.north west, anchor=south west, inner sep=0cm, circle, minimum height=1em, font=\bf] (no4){4};
    \end{tikzpicture}
    }
    \caption{
    The Ag-ODD Framework for deriving the agricultural operational design domain (Ag-ODD). The initial Ag-ODD \protect\circled{2} and its associated logical scenarios \protect\circled{3} are derived from the defined use cases \protect\circled{1}. Once these are established, the iterative verification process \protect\circled{4} begins. During this process, inconsistencies and gaps within the Ag-ODD definition are often exposed by the logical scenarios. Any resulting modifications must then be formalized as a revised Ag-ODD by comparing them to the initial input parameters. This iterative procedure, depicted by the two verification arrows, continues until the Ag-ODD reaches a stable state. Comparing the Ag-ODD with the input parameters: I) functional requirements, II) system capabilities, and III) the results of the Hazard Analysis and Risk Assessment (HARA) as detailed in \protect\cref{fig:Ag-ODD}, as well as the logical scenario derivation as illustrated in \protect\cref{fig:Ag-PEGASUS}.}   \label{fig:overview}
\end{figure*}

The objective identified is to develop the framework in \cref{fig:overview} for an ODD specification within the agricultural context (Ag-ODD Framework). To that end, it synthesizes an approach by integrating elements from multiple existing standards and frameworks. These elements are systematically adapted to align with the current requirements of agricultural industries, providing sufficient flexibility to support integration into existing standards and frameworks. 

The Ag-ODD Framework must be explicitly designed in a manner that allows the modeling of the agricultural operational tasks and processes in whole. Addressing its diverse characteristics introduced with such an extension is the primary benefit of the framework presented in this work.

Consequently, the following sections will deal with a general description of the framework. 
The Ag-ODD Framework is based on a generic ASAM OpenODD model enhanced by aspects of CityGML as well as additional elements to allow for a generic process- and context-sensitive ODD description. For the sake of differentiation, the term Agricultural ODD (Ag-ODD) is being introduced. The term will be used in its now defined short term Ag-ODD.

The adaptations made in this work are: 
\begin{itemize}
    \item A consecutive adaptation of automotive-focused ASAM OpenODD description towards agricultural needs, especially introducing off-road nomenclature. This adaptation allows an easy integration into already existing modeling and simulation frameworks, as well as combined use for on-road and off-road use cases.
    \item An enhancement to the model by introducing a process-related attribute context.
    \item The establishment of a method for context and user-centric abstraction, especially for the sake of modeling and verification processes. For example, different verification strategies, especially when simulation is being used, will demand context-sensitive detailing. Specific properties or attributes may not be relevant to the given verification strategy; thus, the model should allow for a higher level of abstraction. The method being presented is a combination of CityGML's LoD and ASAM OpenODD's attribute property \textit{permissive}~($\cup$)  and \textit{restrictive}~($\cap$). 
\end{itemize}

\section[A Framework for Creating of Agricultural Operational Design Domains]{\texorpdfstring{A Framework for Creating of Agricultural\\ Operational Design Domains}{A Framework for Creating of Agricultural Operational Design Domains}}
The following section will provide an overview of the four core components that constitute the foundation of the Ag-ODD Framework. Each component in \cref{fig:overview} is discussed in detail in the following subsections. The objective of this framework is to ensure maximum flexibility and broad applicability within the agricultural industry. To that end, it is designed to commence with use cases, which are a customary element in the development of agricultural systems. The starting point is the definition of use cases \circled{1}, which describe the intended application and provide the basis for further development. The Ag-ODD \circled{2}, as cab be seen in \cref{fig:Ag-ODD}, is iteratively derived from these use cases, along with the I) functional requirements, II) system
capabilities, and III) the results of the Hazard Analysis and Risk Assessment (HARA) as input parameters.

In addition, as shown in \cref{fig:Ag-PEGASUS}, the Ag-ODD Framework includes scenario creation by using the 7-Layer Model approach, which bases on the 6-Layer Model from PEGASUS. Logical scenarios \circled{3} are applied to the framework to support Ag-ODD verification, and concrete scenarios can ultimately be used to validate the function within the Ag-ODD. In an systematic iterative verification process \circled{4}, illustrated in \cref{fig:iterative_verification}, the Ag-ODD will be shaped by the proposed coherent framework out of all four components \circled{1} to \circled{4}.

\subsection{Term Definitions}
By creating an Ag-ODD Framework, the use of vocabulary defined in existing standards is desired. However, this subsection will clarify the ambiguous definitions of some terms.
\subsubsection{Use Cases \protect\circled{1}}
This work employs the term \textsc{use case} to denote a structured description that captures the function to be realized, the environment in which it operates, and the process in which it is embedded. Initially, the definition of a use case may remain at a high level of abstraction, where exact parameters are not yet fixed. For instance, a generic use case could be formulated as \emph{tilling a wheat stubble field}. The use case is subsequently refined to facilitate the derivation of the Ag-ODD, e.g., \emph{tilling a wheat stubble field in all weather conditions in urban and rural areas in Europe}. This example elucidates the principle, yet it is not exhaustive; additional aspects, such as interaction with other objects, must also be considered. In this work, use cases are employed as the foundation for deriving the input parameters of the Ag-ODD.

\subsubsection{Process}
The term \textsc{process}, in this work, is employed to denote the agricultural process; a specific series of actions and operations that are undertaken for the purpose of cultivating land and raising crops or livestock. It encompasses all the steps involved in food and fiber production, from cultivating soil, planting, raising and harvesting crops, rearing, feeding, and managing animals. The storage or preservation of raw materials prior to the commencement of the production process, in addition to the storage, preservation, handling, or movement of finished goods, is not considered to be within the definition of process. It is noteworthy that the term \textsc{process} is regarded at the framing level rather than the machine level. In other words, the process within a combine harvester, such as the grain flow, is classified as a machine process.

\subsubsection{Agricultural Operational Design Domain}
The term \texttt{Ag-ODD} refers to an Agricultural Operational Design Domain. As shown in \cref{fig:Ag-ODD}, it is a conceptual model that illustrates how an Ag-ODD can be structured and uniquely described. Rather than delineating a discrete Ag-ODD for a given use case, it establishes a concept for describing one.

\subsubsection{7-Layer Model}
The term \texttt{7-Layer model} refers to the model used to derive agricultural logical scenarios. As \cref{fig:Ag-PEGASUS} illustrates, the 7-Layer model is based on the 6-Layer Model of the PEGASUS project. A process layer is added to account for agricultural processes that alter the environment.

\subsubsection{Iterative Verification Process}
In the context of this work, the term \texttt{iterative verification process} refers to the process by which the logical scenarios of the 7-layer model and the Ag-ODD verify each other, as illustrated in \cref{fig:iterative_verification}. Since new scenarios can be discovered or existing ones modified at any time, this process is performed multiple times.

\subsubsection{Agricultural Operational Design Domain Framework}
The term \texttt{Ag-ODD Framework} stands for Agricultural Operational Design Domain Framework. It encompasses the complete methodological process illustrated in \cref{fig:overview}, which outlines the structured sequence of steps required to derive, describe, and verify an Ag-ODD. It begins with formulating the use case, continues with generating logical scenarios using the adapted 7-layer model, and concludes with the iterative verification process that ensures consistency and completeness between the Ag-ODD and its corresponding logical scenarios.

\begin{figure*}[H]
    \centering    
    \begin{tikzpicture}
    \input{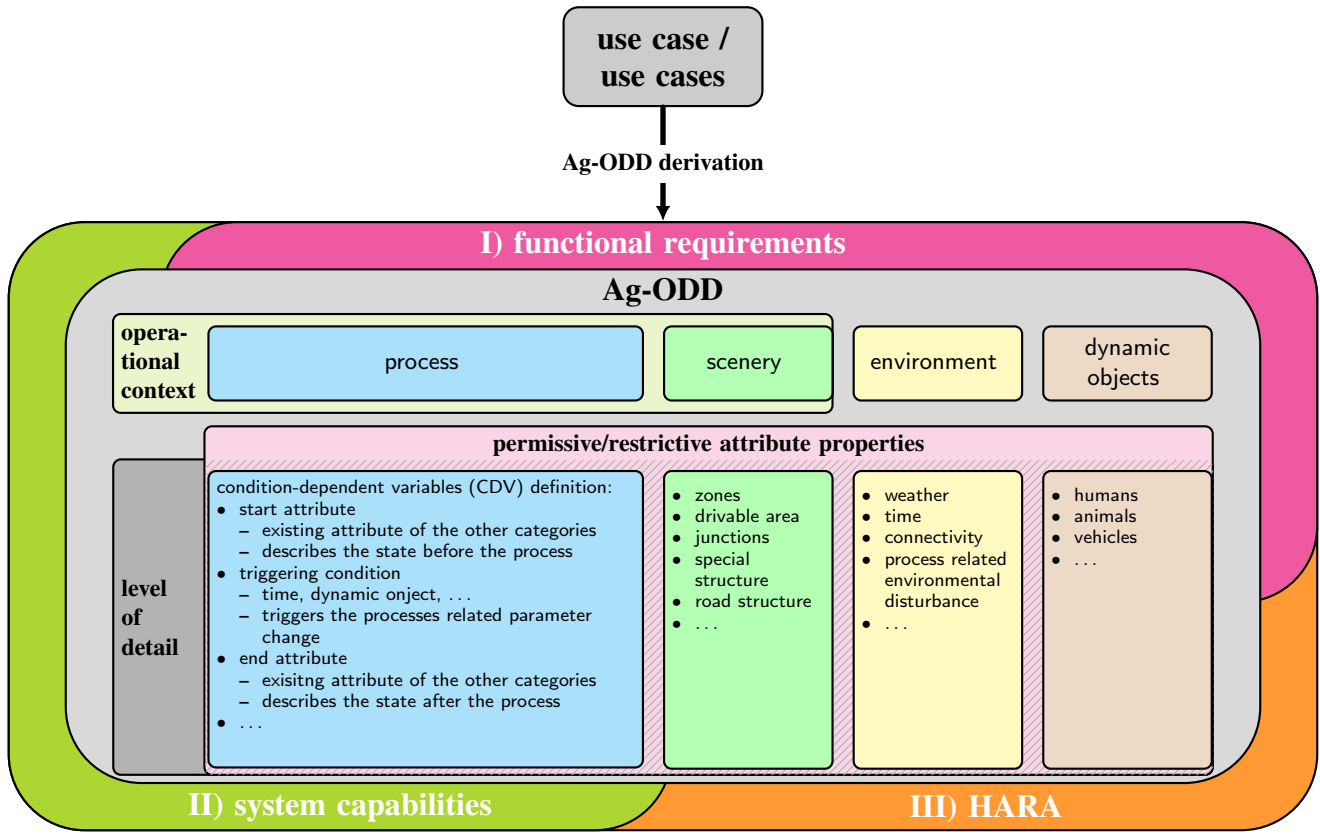}
    \node(uc)[draw, above=1.5cm of hara, thick, fill=black!20,text width=2.4cm, minimum height=4em, align=center, rounded corners=0.2cm, font={\bf \large}] {use case / use cases};
    \draw[draw,line width=2pt, line cap=round, -{Latex[length=2mm, width=2mm]}] (uc)-- node[midway, fill=white, font={\bf}]{Ag-ODD derivation}(hara);
    \end{tikzpicture}
    \caption{
    The Agricultural Operational Design Domain (Ag-ODD) \protect\circled{2} concept. The derivation is done from the following three values: I) functional requirements, II) system capabilities, and III) the results of the Hazard Analysis and Risk Assessment (HARA) as framing limitation. The Ag-ODD is composed of four primary categories: the \protect\boxeded{process}{\processColor}, the \protect\boxeded{scenery}{\sceneryColor}, the \protect\boxeded{environmental}{\environmentColor} condition, and the \protect\boxeded{dynamic objects}{\objectsColor}. In addition, each attribute is defined by a \protect\boxeded{level of detail}{\lodColor} and a \protect\boxeded{permissive/restrictive properties}{\propColor}.  In order to ensure a comprehensive understanding, it is necessary to consider the categories of the operational context sequentially. The process category includes condition-dependent variables. The framing limitations III) HARA are influenced by the I) functional requirements  and II) system capabilities , creating a cyclical dependency between the Ag-ODD and the framing limitations. The starting point for defining Ag-ODD is the use case or use cases \protect\circled{1} in connection with the framing limitations or Ag-ODD itself.}
   \label{fig:Ag-ODD}
\end{figure*}

\subsection[Agricultural Operational Design Domain \protect\circled{2} Derivation]{\texorpdfstring{Agricultural Operational Design Domain \protect\circled{2} \\ Derivation}{Agricultural Operational Design Domain \protect\circled{2} Derivation}}
\label{sec:Ag-ODD_definition}
As one possible initial step, driven by one or several use cases \protect\circled{1}, the derivation of an Ag-ODD involves establishing its conceptual framework and defining its parameters. In other words: How the Ag-ODD should be understood and how the Ag-ODD-Framework should be approached. As discussed beforehand, this work adopts the ASAM OpenODD definition, which builds on the definitions of SAE J3016 and ISO 34503. Accordingly, an Ag-ODD specifies the operating conditions under which a function is designed and engineered to operate safely and reliably. These conditions include environmental factors, infrastructure, geographical features, and dynamic elements.

As shown in \cref{fig:Ag-ODD}, the Ag-ODD is derived from use cases by identifying of I) functional requirements, II) system capabilities, and III) HARA results. These three components, I) to III), are interdependent and form the input space for defining the Ag-ODD. 

By using elements of the ASAM OpenODD, the Ag-ODD's structure uses its description logic, which distinguishes three top-level categories. The most substantial modification concerns the original scenery category, which has been redefined as \boxeded{operational context}{\operationalColor} and subdivided into two parts: \boxeded{process}{\processColor} and \boxeded{scenery}{\sceneryColor}. Where the \boxeded{scenery}{\sceneryColor} corresponds to the ASAM scenery category, which describes static objects and environmental structural features that can be altered with a finite amount of effort, the \boxeded{process}{\processColor} category describes the agricultural process. The \boxeded{environment}{\environmentColor}  category encompasses all weather-related factors that cannot be modified with reasonable effort, such as temperature, sunlight, and cloud coverage. This category also includes influences such as time of day or season, connectivity-related conditions such as network availability and signal stability, and process-related environmental disturbances, such as dust generation, soil movement, and emissions from ongoing agricultural operations. The \boxeded{dynamic objects}{\objectsColor} category continues to cover all potentially moving objects. 

By introducing \textbf{condition-dependent variables (CDV)}, the additional \boxeded{process}{\processColor} category enables the representation of agricultural processes by defining selected predefined attributes. Linking this to a triggering condition and an end attribute allows a process-related description of the parameter change. The triggering condition can be a relative time, a state change, or another measurable event.  In other words, a CDV consists of a start attribute, one or several triggering conditions, and an end attribute without introducing new attributes beyond those in the other categories. For instance, the attribute \textit{stubble field} may change to \textit{cultivated field} when a cultivator passes over it. Similarly, the attribute \textit{crop height} may change from \SI{70}{\centi\meter} to \SI{20}{\centi\meter} when interacting with a cutter bar. As differing process-related parameter changes may occur during the same operation, attributes can be linked to multiple CDV; for instance, one CDV could represent harvesting, while another could represent lodged grain.

By adapting to the Ag-ODD needs and improving its practical usability, two concepts are integrated in addition: the \boxeded{permissive/restrictive attribute properties}{\propColor} from ASAM OpenODD and the \boxeded{level of detail (LoD)}{\lodColor} concept from CityGML. Together, these concepts provide a powerful, matrix-like mechanism that maintains the precision and efficiency of the Ag-ODD description.

In consequence, each  Ag-ODD attribute can be designated as either permissive ($\cup$) or restrictive ($\cap$). A permissive attribute implies the inclusion of all instances of the attribute that are not explicitly listed. In contrast, a restrictive attribute requires the explicit enumeration of all valid instances. For instance, if the \textit{tractor} attribute is permissive, all tractors, regardless of their shape, size, or color, are included in the Ag-ODD. If marked restrictive, only those tractors explicitly defined are considered part of the Ag-ODD, and if none are specified, no tractors are included at all. However, this restrictive approach may result in an impractically high level of specification effort. To mitigate this issue, the concept of LoD is intended.

\begin{figure*}[H]
    \centering
    \begin{tikzpicture}[
            field/.style={trapezium, trapezium left angle=60, trapezium right angle=120, draw, fill={rgb,255:red,71; green,82; blue,60}, text width=2.5cm, text height=1.5cm,inner sep=0cm, postaction={pattern={Lines[angle=60,distance=1.2mm,  line width=0.6mm]},pattern color={rgb,255:red,119; green,124; blue,90}},},
            fieldT/.style={text width=6.cm, font=\footnotesize,},
            fieldO/.style={draw, thick, fieldoutline, text width=2.0cm, text height=1.5cm,inner sep=0cm,},
            fieldS/.style={trapezium, trapezium left angle=60, trapezium right angle=120, text width=0mm, text height=1.5cm, inner xsep=-2.5mm, inner ysep=0mm,fill=gray},
            ]
    \input{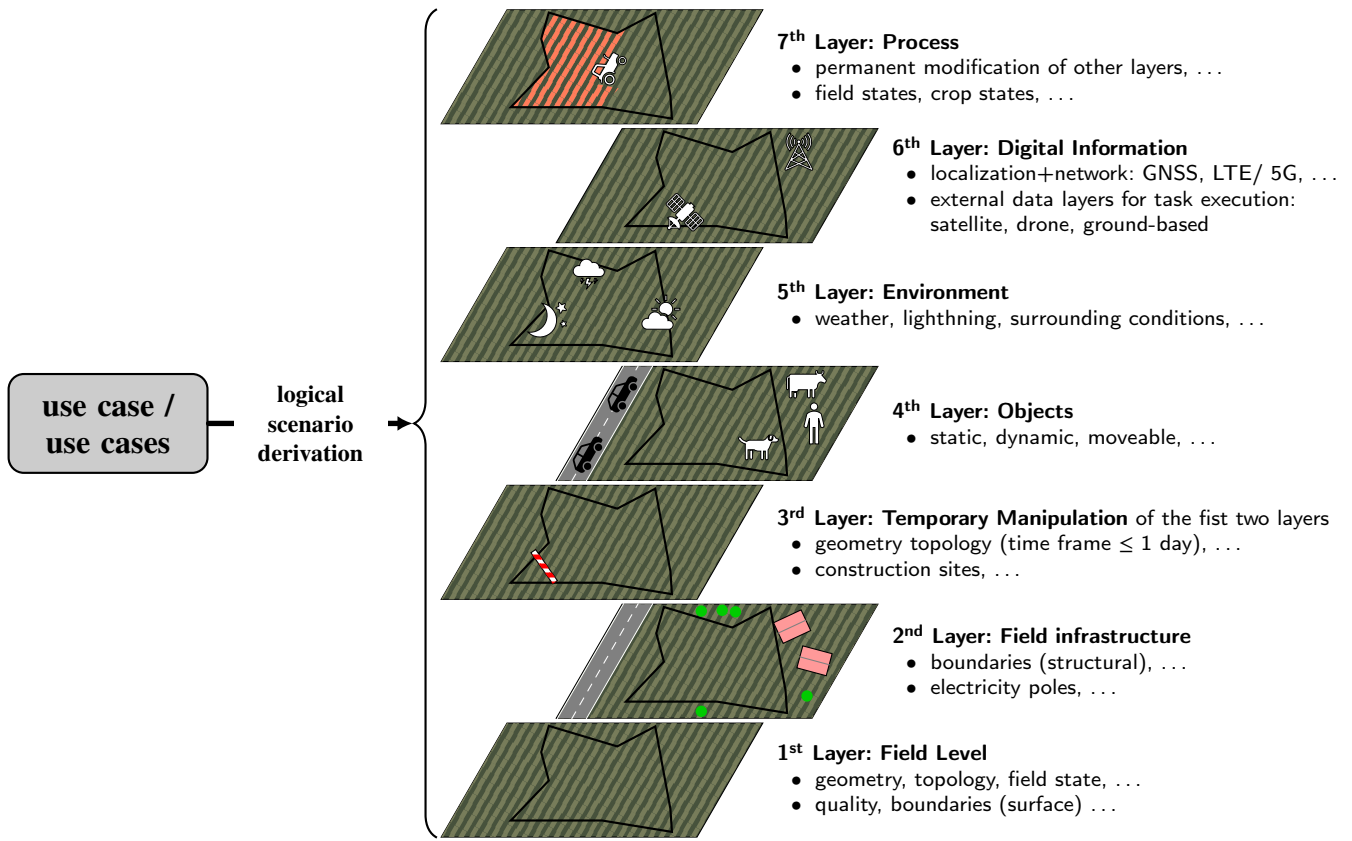}
    \node(uc)[draw, left=5.0cm of field4, thick, fill=black!20,text width=2.4cm, minimum height=4em, align=center, rounded corners=0.2cm, font={\bf \large}] {use case / use cases};
    \draw[draw,line width=2pt, line cap=round, -{Latex[length=2mm, width=2mm]}] (uc)-- node[midway, fill=white, text width=1.8cm, align=center, font={\bf}]{logical scenario derivation}($(sumScenario)$);
    \end{tikzpicture}
    \caption{The 7-Layer Model is designed for use in agricultural scenarios and can be used to derive logical scenarios, including agricultural processes from use cases. The 7-Layer Model comprises the six PEGASUS layers \protect\citep{PPO2021} with slight modifications, as well as a 7\textsuperscript{th} process layer that can potentially alter the other layers through process descriptions.}
   \label{fig:Ag-PEGASUS}
\end{figure*}

The LoD concept allows attributes to be refined in a structured way, avoiding excessive complexity. In general, a lower LoD inherently indicates a higher level of abstraction. Continuing with the tractor example, a permissive tractor attribute includes all tractors. A restrictive attribute, on the other hand, could be specified by the LoD to include only green tractors. The newly defined sub-attribute, \textit{green tractors}, inherits a permissive property, meaning all green tractors are included unless constrained by a more restrictive LoD. For example, one could further restrict the LoD to \textit{green tractors under \SI{200}{\kilo\watt}}, thereby narrowing the scope. Even if an attribute is further specified so that it has a restrictive limit, e.g., such as \SI{200}{\kilo\watt}, it remains permissive. Only when it is unambiguous does it become a restrictive attribute. Thus, LoD enables the precise yet efficient specification of which elements belong to the Ag-ODD. 

To avoid discrepancies, treat the entire Ag-ODD as restrictive by default, and treat each mentioned attribute as permissive. This default means that any unmentioned attribute is not part of the Ag-ODD. However, as soon as an attribute is mentioned, all of its unmentioned sub-attributes are included in the Ag-ODD by default.

\Cref{sec:ex_Ag-ODD} illustrates the application of these concepts through practical examples. It demonstrates how permissive/restrictive attribute properties and the LoD contribute to a structured yet manageable Ag-ODD definition. In summary, the Ag-ODD extends the conventional ODD framework in two essential ways: first, by introducing a \textit{Process} field that allows the description of processes and state transitions; and second, by incorporating \boxeded{permissive/restrictive attribute properties}{\propColor} and \boxeded{LoD}{\lodColor} refinements to enable unambiguous and efficient specification.

\subsection{Logical Scenario \protect\circled{3} Derivation}
\label{sec:logical_scenario}
As an other possible initial step, the logical scenarios must first be derived from the previously defined use cases~\circled{1}. These scenarios are abstract descriptions that do not yet contain fixed parameter values, allowing them to encompass a wide range of potential concrete scenarios. 
The Ag-ODD Framework proposed in \cref{fig:overview}, builds on the scenario description model from the PEGASUS project to generate such logical scenarios efficiently. However, the 7-Layer Model is not the only possible approach. Any systematic method of generating equivalent scenarios could be used to achieve the same goal. The 7-Layer Model was chosen for the Ag-ODD Framework primarily because it is widely used in industry and research and can be easily adapted for agricultural applications by adding a process layer. 

As illustrated in \cref{fig:Ag-PEGASUS}, the original PEGASUS model is extended by this additional process layer, which ensures that process-related aspects are not lost in the description. The 1\textsuperscript{st} layer of the model describe fundamental aspects, such as geometry, topology, and state. The 2\textsuperscript{nd} layer adds infrastructure features, and the 3\textsuperscript{th} layer accounts for temporary modifications to the two lower layers. For example, this could include construction sites located on or adjacent to a field. The 4\textsuperscript{th} layer specifies all objects relevant to the scenario, and the 5\textsuperscript{th} layer covers environmental factors such as lighting and weather conditions. As before, the 6\textsuperscript{th} layer describes the digital information in the logical scenario. The newly introduced 7\textsuperscript{th} process layer allows us to describe permanent modifications to the other layers, thereby capturing processes that actively alter the operating environment.

Structuring scenario descriptions in this manner enables the development of a comprehensive set of logical scenarios, ensuring that all relevant concrete scenarios are covered. These logical scenarios provide the foundation for verifying the Ag-ODD once its initial specification is complete.

\begin{figure*}[H]
    \centering
    \begin{tikzpicture}[
    hexagon/.style={regular polygon, regular polygon sides=6, draw=none, fill={rgb,255:red,21; green,95; blue,130}, inner sep=0cm,text width=2.4cm, minimum size=2.4cm, font=\scriptsize, align=center},
    hexagonODD/.style={regular polygon,regular polygon sides=6,draw, fill={rgb,255:red,224; green,224; blue,224},inner sep=0cm, text width=0.7cm, minimum size=0.7cm, align=center, font={\bf \tiny}},
    scenario/.style={regular polygon,regular polygon sides=6,draw, fill={rgb,255:red,193; green,228; blue,245},inner sep=0cm, text width=0.85cm, minimum size=0.85cm, align=center, font={\scriptsize \ttfamily}},
    rl/.style={draw=purple,line width=2pt, line cap=round},
    gl/.style={draw=lime!80!black,line width=2pt, line cap=round, dashed},
    optl/.style={draw,line width=2pt, line cap=round, -{Latex[length=2mm, width=2mm]},shorten <=0.2cm, , shorten >=0.2cm},
    ]
    \input{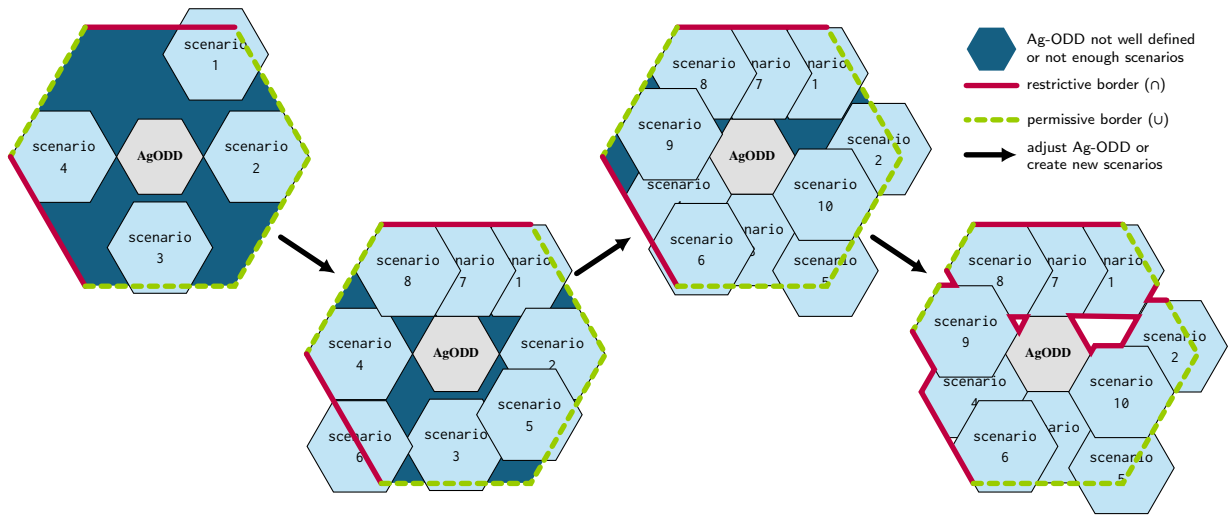}
    \begin{scope}[scale=0.8, every node/.append style={transform shape}]
    \node(lcase)[hexagon, right= 2.5cm of case3.corner 1, anchor=corner 2, text width=0.5cm, minimum size=0.5cm] {};
    \node(lcaseText)[right= 0.05cm of lcase, text width=3.2cm, minimum height= 2em, font=\scriptsize]{Ag-ODD not well defined\\ or not enough scenarios};
    \node(perbText)[below= -0.1cm of lcaseText, text width=3.2cm, minimum height= 2em, font=\scriptsize]{restrictive border ($\cap$)};
    \node(resbText)[below= -0.1cm of perbText, text width=3.2cm, minimum height= 2em, font=\scriptsize]{permissive border ($\cup$)};
    \node(adjustText)[below= -0.1cm of resbText, text width=3.2cm, minimum height= 2em, font=\scriptsize]{adjust Ag-ODD or\\ create new scenarios};
    \draw[rl] (perbText-|lcase.corner 6)--(perbText-|lcase.corner 3);
    \draw[gl] (resbText-|lcase.corner 6)--(resbText-|lcase.corner 3);
    \draw[optl,shorten <=0cm, , shorten >=0cm] (adjustText-|lcase.corner 3)--(adjustText-|lcase.corner 6);
    \end{scope}

    \end{tikzpicture}
    \caption{Iterative verification process \protect\circled{4} between the Ag-ODD \protect\circled{2} and the logical scenarios \protect\circled{3}. A predefined Ag-ODD, as drawn as dark blue hexagon, means that the Ag-ODD is not yet well enough defined or that there are not yet enough scenarios. As soon as the scenarios cover the entire Ag-ODD, the Ag-ODD \protect\circled{2} is verified against the scenarios \protect\circled{3} and vice versa. A green dashed border means that this boundary is permissive ($\cup$)---everything that is not explicitly mentioned is still included; a red border means that this boundary is restrictive ($\cap$)---everything that is not explicitly mentioned is excluded. In the first two iterations, additional \texttt{scenarios} are added to better cover the Ag-ODD. Within the third iteration, the Ag-ODD is adjusted so that no further scenarios are need or because these parts of the Ag-ODD cannot be supported by framing limitations.}
   \label{fig:iterative_verification}
\end{figure*}

\subsection[Agricultural Operational Design Domain Verification \protect\circled{4}]{\texorpdfstring{Agricultural Operational Design Domain\\ Verification \protect\circled{4}}{Agricultural Operational Design Domain Verification \protect\circled{4}}}
\label{sec:verification}
Verifying the Ag-ODD within the Ag-ODD Framework is a crucial step toward enabling reliable autonomous functions. The purpose of verification is to identify potential gaps in the Ag-ODD definition \circled{2} and edge cases that should not or cannot be included, e.g., due to insufficient system capabilities. 

Consequently, this section describes an iterative process intended to identify inconsistencies and close gaps within the Ag-ODD definition. This is done via an iterative comparison between the Ag-ODD \circled{2} on the one hand and the scenario definition \circled{3}, e.g., via the enhanced 7-Layer Model, on the other hand. The comparison includes verification of the \boxeded{permissive/restrictive attribute properties}{\propColor} and is facilitated by incorporating the \boxeded{LoD}{\lodColor} concept. The iterative verification process \circled{4} of the Ag-ODD Framework is detailed in \cref{fig:iterative_verification}.

\begin{description}
    \item[\textbf{Initial State: }]
    First, an Ag-ODD \circled{2} must be specified, and an initial set of logical scenarios \circled{3} must be available to begin the iterative verification process \circled{4}. This predefined Ag-ODD, illustrated by the dark blue hexagon in  \cref{fig:iterative_verification}, needs to be aligned with logical scenarios, illustrated by light blue hexagons.  The green and red edges symbolize permissive and restrictive attributes, respectively. Borders should be considered with caution as soon as they become permissive. Since they combine all subordinate information and more specific attributes, the scenarios themselves contain a multidimensional array of attributes and parameters, which could result in a very large number of test cases. During function validation, it is important to ensure that permissive borders are adequately tested. Note, \cref{fig:iterative_verification} is a metaphor, as a real Ag-ODD can be multidimensional and contain significantly more parameters and relationships.
    \item[\textbf{First Iteration:}] The first iteration reveals that the initial Ag-ODD is covered by too few logical scenarios, leaving more than half of its parameter space unverified. Additionally, \texttt{scenario 1} exceeds a restrictive $\cap$ boundary, meaning a restrictive attribute assumes a value not permitted by the Ag-ODD. Conversely, transgressions across green, permissive $\cup$, edges are acceptable because permissive attributes allow additional values by definition. This analysis indicates that two things can be done. Either the Ag-ODD itself must be adapted, or, preferably, further logical scenarios should be defined.
    \item[\textbf{Consecutive Iterations:}]
    A significantly larger set of scenarios is added in the second iteration. \texttt{scenario 1} is adjusted so that it no longer exceeds a restrictive limit; however, the newly introduced \texttt{scenario 6} now does. Additionally, parts of the Ag-ODD parameter space remain uncovered. In the third iteration, further logical scenarios were created to expand coverage, but a  restrictive boundary is still crossed by \texttt{scenario 6}.
    \item[\textbf{Final Iteration:}]
    In the final iteration, all logical scenarios are refined to maximize coverage of the Ag-ODD parameter space without exceeding restrictive boundaries. At the same time, parameter spaces for which no valid scenarios exist, such as those that exceed the system's capabilities, are removed from the Ag-ODD definition. This results in the creation of explicit restrictive $\cap$ limits. Attributes adjacent to these spaces must then be assigned the restrictive property within the Ag-ODD.
\end{description}

Each iteration, particularly those involving modifications to the Ag-ODD \circled{2} or the set of logical scenarios \circled{3}, requires a subsequent review of the framing limitations I) functional requirements, II) system capabilities, and III) HARA results to ensure complete consistency with the defined use cases \circled{1}. If discrepancies arise, the use cases must be adjusted or further corrective actions must be considered.

\subsection{Framework Integration}
The Ag-ODD Framework presented here is designed to be generically applicable, as it builds on and integrates established ODD concepts. Many of these concepts are already part of standard development and validation processes. Consequently, the framework can be adapted with adequate effort while enabling a consistent description of Ag-ODD for agricultural applications. As illustrated in \cref{fig:overview}, both strands of the framework are indispensable; regardless of which was initially emphasized, a complete and verified Ag-ODD can only be achieved through their combination.

\cref{sec:examples} further exemplifies the framework by illustration two  break-down structures for representative use cases, which enables the tracing of the entire process from the initial definition to a fully specified and verified Ag-ODD. Furthermore, \cref{sec:val_ex_fun} demonstrates the framework's practical applicability by providing examples of how an Ag-ODD can validate the functions of autonomous agricultural machinery.

\begin{table*}[tbh]
\centering
\renewcommand{\arraystretch}{1.2}
\caption{The Ag-ODD for the cultivation use case is shown below. The \protect\boxeded{process}{\processColor} category generates condition-dependent variables (CDV) with start attribute SA1, condition C1, and end attribute EA1. The descriptions of the \protect\boxeded{scenery}{\sceneryColor}, the \protect\boxeded{environment}{\environmentColor} and the \protect\boxeded{dynamic objects}{\objectsColor} categories are represented with different \protect\boxeded{levels of detail (LoD)}{\lodColor}. The first iteration is highlighted with white, the second with blue, and the third with red cell background. The \protect\boxeded{permissive/restrictive attribute properties}{\propColor}, shows whether an attribute is permissive ($\cup$) or restrictive ($\cap$) is in the Type column. The Ag-ODD is by definition, entirely restrictive---anything not listed is excluded. The current or final Ag-ODD is represented from the right by the dashed line.
}
\scalebox{0.84}{
\begin{NiceTabular}
{>{\raggedright\arraybackslash}p{0.25\textwidth} 
>{\centering\arraybackslash}p{0.03\textwidth} 
>{\centering\arraybackslash}p{0.05\textwidth} 
>{\raggedright\arraybackslash}p{0.25\textwidth} 
>{\centering\arraybackslash}p{0.03\textwidth} 
>{\centering\arraybackslash}p{0.05\textwidth} 
>{\raggedright\arraybackslash}p{0.2\textwidth} 
>{\centering\arraybackslash}p{0.03\textwidth} 
>{\centering\arraybackslash}p{0.05\textwidth} 
}
\boxeded{LoD0}{\lodColor}                                      & \textbf{}          & \boxeded{Type}{\propColor}          & \multicolumn{1}{||l}{\boxeded{LoD1}{\lodColor}}                                                  &  & \boxeded{Type}{\propColor}                & \multicolumn{1}{||l}{\boxeded{LoD2}{\lodColor}} &  & \boxeded{Type}{\propColor}  \\ \hline
\boxeded{scenery}{\sceneryColor}                                   & \textbf{}          & \textbf{$\mathbf{\cap}$}          & \multicolumn{1}{||l}{}                                                               &  &                             &  \multicolumn{1}{||l}{}      &  &               \\ 
                                                   &                    &                       & \multicolumn{1}{||l}{\cellcolor[HTML]{FFCCC9}Fields in GER  }                        & \cellcolor[HTML]{FFCCC9} & \cellcolor[HTML]{FFCCC9}$\cup$& \multicolumn{1}{||l}{}  &  &              \\ 
\multirow{-2}{*}{Fields in Europe}                 & \multirow{-2}{*}{} & \multirow{-2}{*}{$\cap$} & \multicolumn{1}{||l}{\cellcolor[HTML]{CBCEFB}Slope $\leq$~\SI{10}{\percent}}                      & \cellcolor[HTML]{CBCEFB} & \cellcolor[HTML]{CBCEFB}$\cup$& \multicolumn{1}{||l}{}                &  &               \\
Crop stubbles ($\leq$~\SI{15}{\centi\meter})                        & SA1                & $\cap$                  & \multicolumn{1}{||l}{}                                                               &  &                             &   \multicolumn{1}{||l}{}              &  &               \\ 
Uneven soil ($\leq$~\SI{15}{\centi\meter})                          & EA1                & $\cap$                  & \multicolumn{1}{||l}{\cellcolor[HTML]{CBCEFB}No lying snow}                         & \cellcolor[HTML]{CBCEFB} & \cellcolor[HTML]{CBCEFB}$\cup$&\multicolumn{1}{||l}{}                 &  &               \\ \hline
\boxeded{environment}{\environmentColor}                               &                    & \textbf{$\mathbf{\cap}$}          & \multicolumn{1}{||l}{}                                                                 &  &                             &   \multicolumn{1}{||l}{}              &  &               \\ 
                                                   &                    &                       & \multicolumn{1}{||l}{\cellcolor[HTML]{FFCCC9}No fog (visibility $\leq$~\SI{50}{\meter})}        & \cellcolor[HTML]{FFCCC9} & \cellcolor[HTML]{FFCCC9}$\cup$&  \multicolumn{1}{||l}{}               &  &               \\  
\multirow{-2}{*}{Conditions without precipitation} & \multirow{-2}{*}{} & \multirow{-2}{*}{$\cap$} & \multicolumn{1}{||l}{\cellcolor[HTML]{FFCCC9}No dust (visibility $\leq$~\SI{50}{\meter})}       & \cellcolor[HTML]{FFCCC9}  & \cellcolor[HTML]{FFCCC9}$\cup$&    \multicolumn{1}{||l}{}             &  &               \\ \hline
\boxeded{dynamic objects}{\objectsColor}                                   &                    & \textbf{$\mathbf{\cap}$}          & \multicolumn{1}{||l}{}                                                                 &  &                             &        \multicolumn{1}{||l}{}         &  &               \\ 
Humans                                             &                    & $\cup$                  & \multicolumn{1}{||l}{\cellcolor[HTML]{CBCEFB}No humans $\geq$~\SI{2}{\meter}}                       &  \cellcolor[HTML]{CBCEFB} & \cellcolor[HTML]{CBCEFB}$\cup$&    \multicolumn{1}{||l}{}             &  &               \\ 
Ego-vehicle                                        &                    & $\cup$                  & \multicolumn{1}{||l}{\cellcolor[HTML]{CBCEFB}Traktor X}                             &  \cellcolor[HTML]{CBCEFB} & \cellcolor[HTML]{CBCEFB}$\cap$&      \multicolumn{1}{||l}{}           &  &               \\ 
Cultivation implement                              & C1                 & $\cup$                  & \multicolumn{1}{||l}{\cellcolor[HTML]{CBCEFB}Width $\leq$~\SI{50}{\meter}}                     &  \cellcolor[HTML]{CBCEFB} & \cellcolor[HTML]{CBCEFB}$\cup$& \multicolumn{1}{||l}{\cellcolor[HTML]{FFCCC9} Implement Y}   & \cellcolor[HTML]{FFCCC9} &\cellcolor[HTML]{FFCCC9} $\cap$      \\ \hline
 & & & & & & & & \\[-2ex]  
\multicolumn{3}{l}{\boxeded{process}{\processColor}}                                                                     & \multicolumn{2}{c}{\textbf{Start}}                                         & \multicolumn{3}{c}{\textbf{Condition}}                  & \textbf{End}         \\ \hline
\multicolumn{3}{l}{24/7 autonomous cultivation (depth limit 15 cm)}                             & \multicolumn{2}{c}{SA1}                                           & \multicolumn{3}{c}{Interaction with C1}        & EA1          

\CodeAfter
  \tikz \draw [line width=2pt, black, dash dot dot] ($(3-|7)+(0.2,0)$) |- (3-|4) |- (3-|4) |- (5-|1) |- (6-|1) |- (6-|4) |- (14-|4) |- (14-|last) |- (13-|last) |- ($(13-|7)+(0.2,0)$) -- cycle; 
\end{NiceTabular}
}

\label{tab:cultivation}
\end{table*}

\section{Evaluation and Examples of Using the Agricultural Operational Design Domain Framework}
\label{sec:ex_Ag-ODD}
This section examines two representative use cases from agricultural practice to illustrate the general applicability and flexibility of the Ag-ODD Framework introduced in this work. Each use case emphasizes different aspects of autonomous field operations and demonstrates how the Ag-ODD \circled{2} can be defined, structured, and verified for a set of specific I) functional requirements as framing limitation.

\subsection{Applications Examples}\label{sec:examples}
The first use case \circled{1} focuses on cultivation, reflecting the ongoing efforts of machinery and implement manufacturers to enable autonomous light soil tillage. This example emphasizes the integration of soil, weather, and process parameters, as well as the interaction between \boxeded{operational context}{\operationalColor} with its \boxeded{process}{\processColor} and \boxeded{scenery}{\sceneryColor} descriptions interact within the Ag-ODD \circled{2}.

The second use case is wheat harvesting, one of the most complex agricultural processes involving dynamic interactions. This use case is used to demonstrate how an Ag-ODD can represent continuously changing process states (e.g., transitioning from a standing crop to a stubble field) and how these states are used to define condition-dependent variables within the operational context.

These use cases \circled{1} show that the Ag-ODD Framework can address current and emerging I) functional requirements.
The examples presented here serve as generic guidance for the application and verification of an Ag-ODD. The examples are not intended to be exhaustive; each framework user should construct and verify the relevant Ag-ODD accordingly.
Further explanations on how to read and interpret the resulting Ag-ODD tables can be found in \cref{sec:general_considerations}.

To facilitate the interpretation of the formal Ag-ODD parametrization the authors are using \emph{lingual descriptions}, which allows as well a harmonized extraction of parameters based on generic user stories, e.g., included in a product description.

\subsubsection{Use Case: Cultivation}
\label{sec:ex_cultivation}
Since the use case was initially defined only in general terms, as \emph{cultivating}, it is specified further in this section to enable its structured translation into an Ag-ODD. At this stage, the I) functional requirements, II) system's capabilities, and III) HARA are not considered in detail because these components emerge from the interaction between the defined use cases \circled{1}, the Ag-ODD \circled{2} description, and the corresponding logical scenarios \circled{3}. The precise determination of those aspects is implied for the Ag-ODD Framework's use, but is beyond the scope of this publication.

The refined use case can be summarized as follows: \emph{Autonomous cultivation operation with continuous 24/7 availability, provided that no precipitation: rain, snow, hail, etc. occurs. The machine operates on fields with stubble up to \SI{15}{\centi\meter} in height, regardless of crop type. Human detection must be ensured, whereas detection of other objects is not required. Soil type is not restricted, and cultivation depth is limited to \SI{15}{\centi\meter}. Operations take place within Europe.}

Clearly, this initial delimitation of the use case is insufficient to derive an unambiguous and complete Ag-ODD. Nevertheless, this initial iteration serves as the starting point for the structured development and subsequent refinement of the Ag-ODD framework.

In the following, the described use case is transferred into the structured Ag-ODD format. The initial classification and assignment of relevant attributes to their respective top-level Ag-ODD categories are summarized in \cref{tab:cultivation}.

Within the tabular visualization, both the iterative procedure and the categories of the target Ag-ODD are depicted. The cells refer either to an attribute or a property. Rows are used to cluster the parameters into the predefined categories \boxeded{process}{\processColor}, \boxeded{scenery}{\sceneryColor}, \boxeded{environment}{\environmentColor} and \boxeded{dynamic objects}{\objectsColor}. In this example the \boxeded{LoD}{\lodColor} method is used to facilitate iterative verification, enabling the parameters to become more specific and precise at each step. The white cells represent the first iteration of verification, the blue cells the second and the red cells the third. Note that this method is not limited to the presented functionality and can also be used in simulation frameworks, although this is beyond the scope of this publication.

In total, \cref{tab:cultivation} shows the finalized Ag-ODD after two iterations. In the iteration described here, only the white cells are considered. To verify the Ag-ODD \circled{2}, logical scenarios \circled{3} must be created. This can be done using the procedure described in \cref{sec:logical_scenario}. To illustrate this procedure, four logical scenarios are provided lingually in form of simple sentences. For the sake of illustration the layer number of \cref{fig:Ag-PEGASUS} is provided in brackets.

\begin{figure*}[H]
    \centering
    \begin{subfigure}[b]{0.15\textwidth}
        \includegraphics[width=\textwidth]{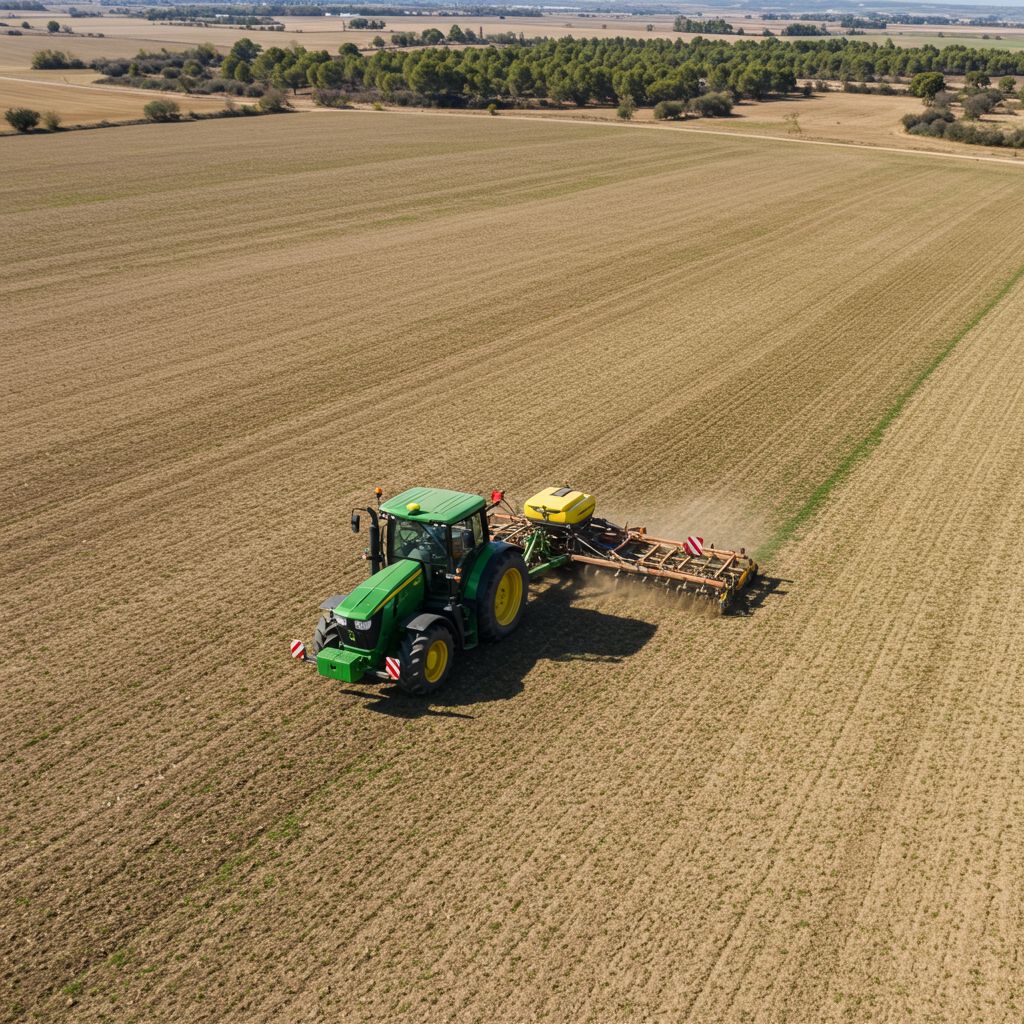}
        \caption{Scenario 1}
        \label{fig:cultivate:sub1}
    \end{subfigure}\hfill
    \begin{subfigure}[b]{0.15\textwidth}
        \includegraphics[width=\textwidth]{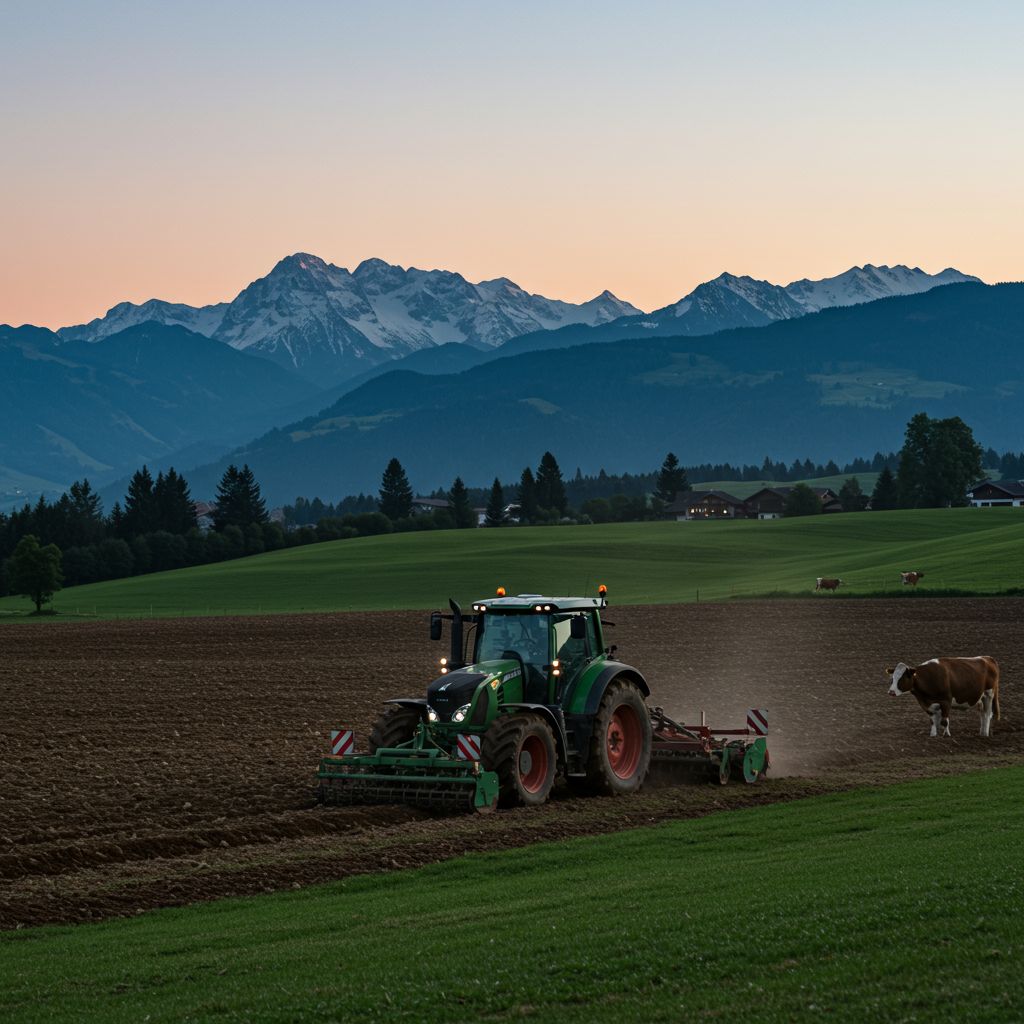}
        \caption{Scenario 2}
        \label{fig:cultivate:sub2}
    \end{subfigure}\hfill
    \begin{subfigure}[b]{0.15\textwidth}
        \includegraphics[width=\textwidth]{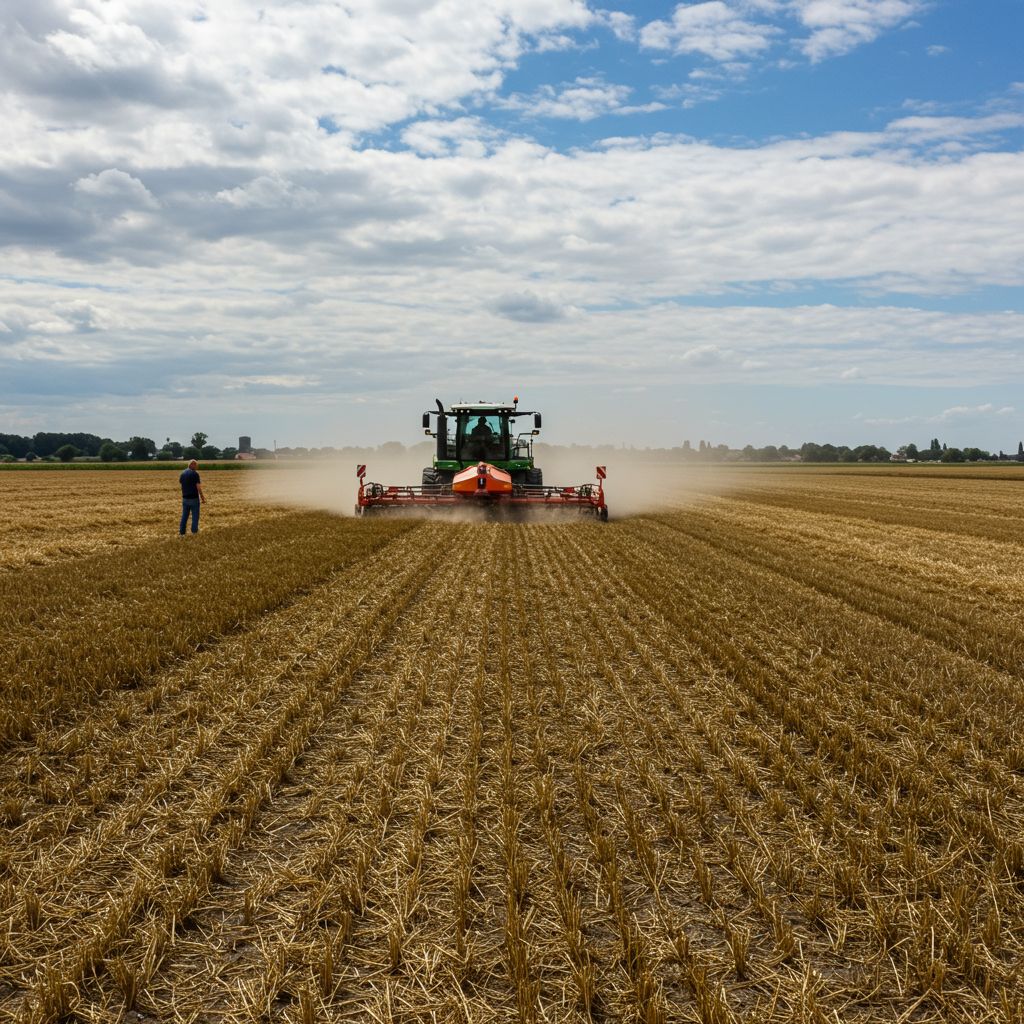}
        \caption{Scenario 3}
        \label{fig:cultivate:sub3}
    \end{subfigure}\hfill
    \begin{subfigure}[b]{0.15\textwidth}
        \includegraphics[width=\textwidth]{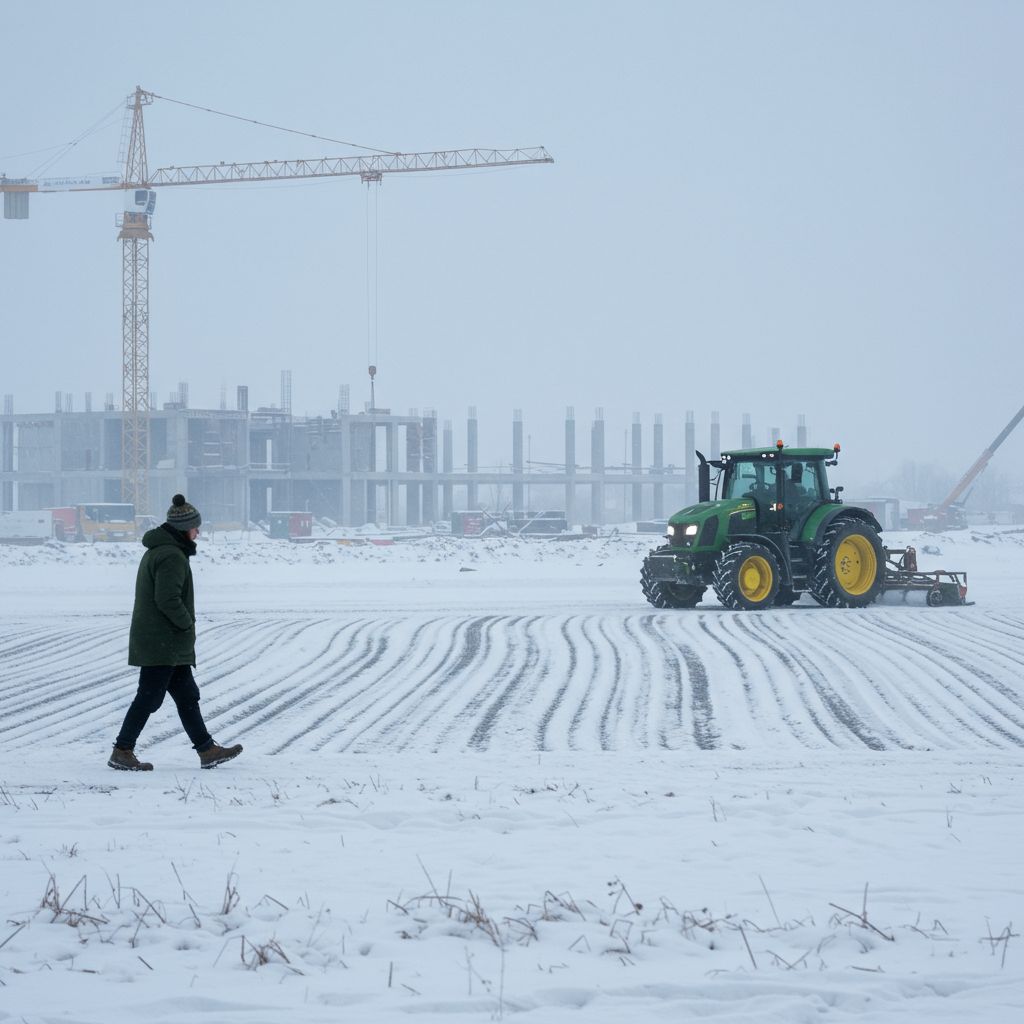}
        \caption{Scenario 4}
        \label{fig:cultivate:sub4}
    \end{subfigure}\hfill
    \begin{subfigure}[b]{0.15\textwidth}
        \includegraphics[width=\textwidth]{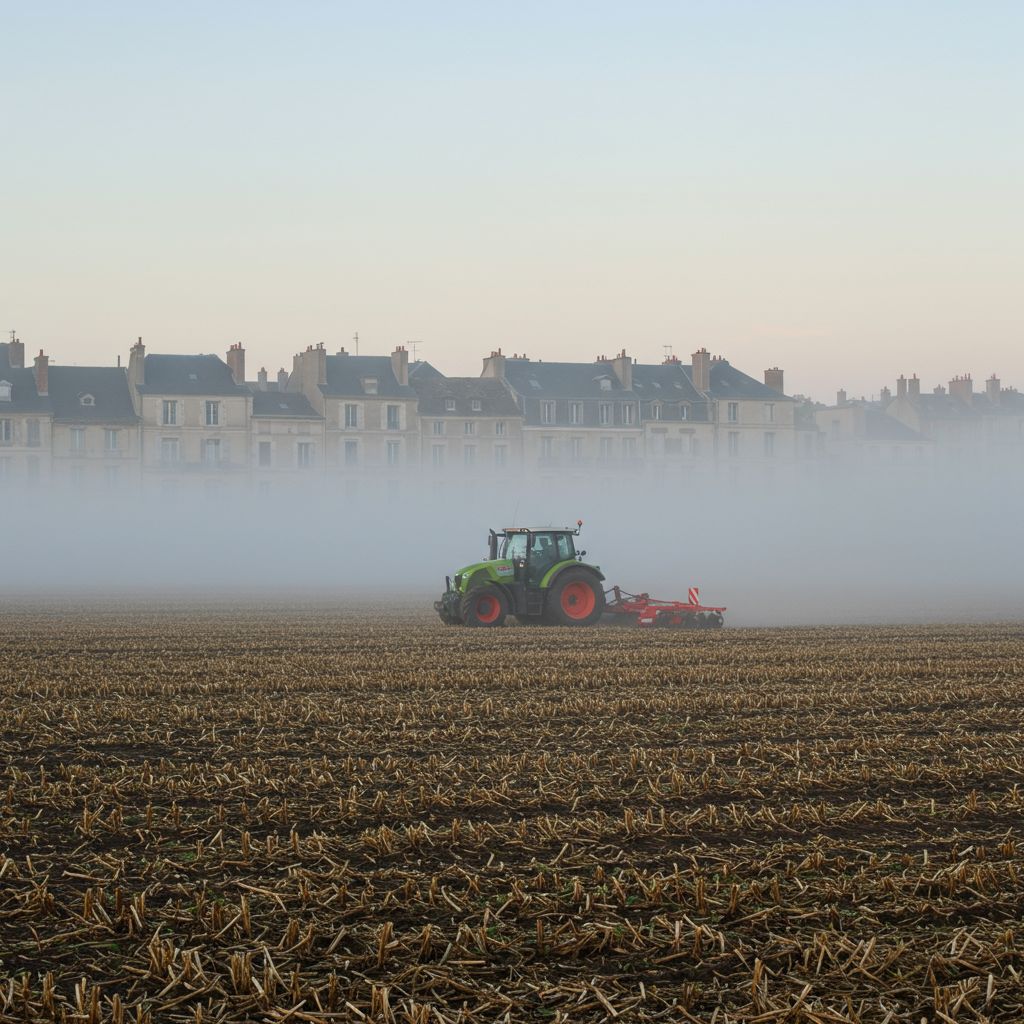}
        \caption{Scenario 5}
        \label{fig:cultivate:sub5}
    \end{subfigure}\hfill
    \begin{subfigure}[b]{0.15\textwidth}
        \includegraphics[width=\textwidth]{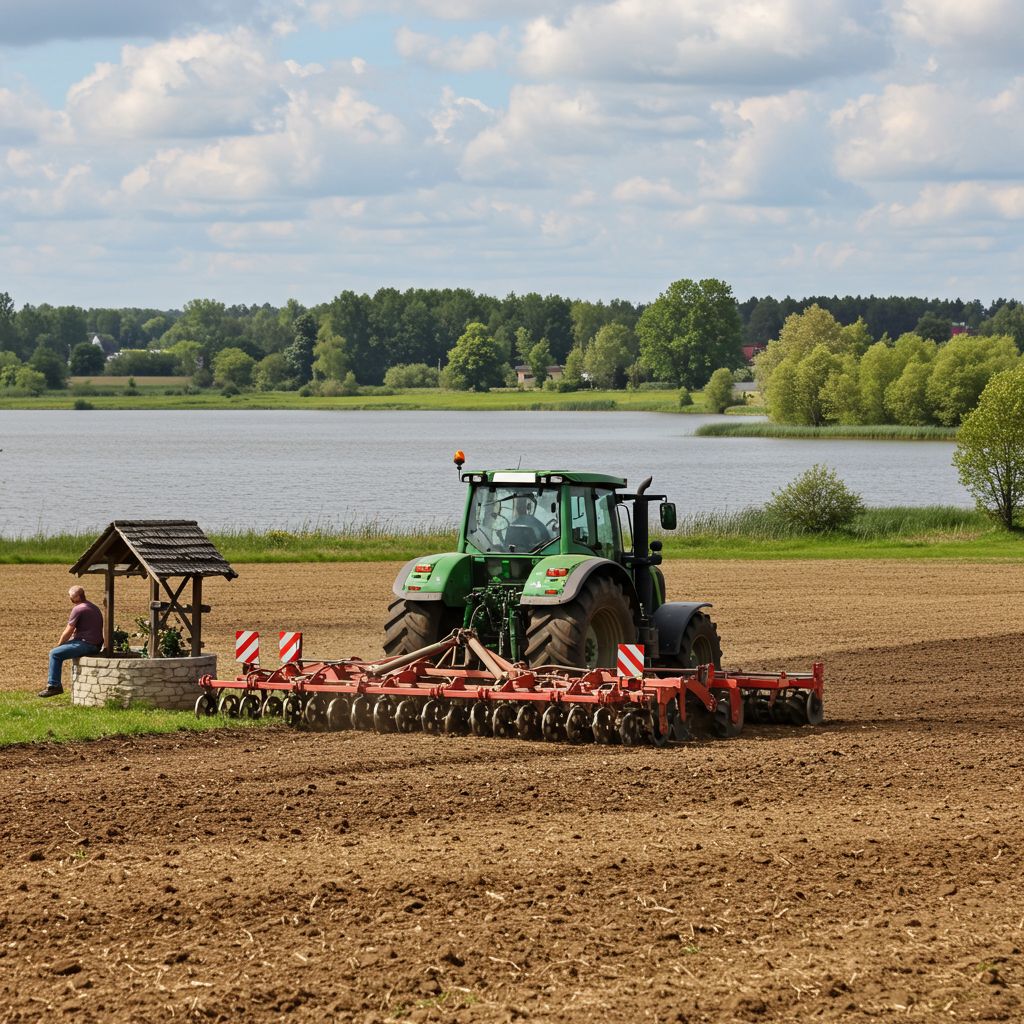}
        \caption{Scenario 6}
        \label{fig:cultivate:sub6}
    \end{subfigure}

    \caption{Logical scenarios \protect\circled{3} derived using the 7-Layer Model are presented here, in the example use case of \protect\emph{cultivating}; these six different visualized scenarios are used during interactive verification \protect\circled{4}. The visualizations (\protect\subref{fig:cultivate:sub1}) to (\protect\subref{fig:cultivate:sub6}) are AI-generated images.}
    \label{fig:cultivation_images}
\end{figure*}

\begin{enumerate}[label=\arabic*., ref=\arabic*]
    \item \label{itm:cult1} In France, an autonomous John Deere tractor is cultivating a rectangular field using a cultivator (1\textsuperscript{st}, 4\textsuperscript{th}, 7\textsuperscript{th} layer) as visulized in \cref{fig:cultivate:sub1}. The area is in a rural region with favorable weather conditions (2\textsuperscript{nd}, 5\textsuperscript{th} layer). Several forested areas are around the field (2\textsuperscript{nd} layer).
    \item \label{itm:cult2} A tractor, in \cref{fig:cultivate:sub2}, is cultivating a field in the foothills of $\leq$\SI{10}{\percent} of the Austrian Alps with an implement (1\textsuperscript{st}, 2\textsuperscript{nd}, 4\textsuperscript{th}, 7\textsuperscript{th} layer). It is dusk, and a cow stands at the edge of the field (4\textsuperscript{th}, 5\textsuperscript{th} layer).
    \item \label{itm:cult3} At midday, like in \cref{fig:cultivate:sub3}, an autonomous tractor is cultivating a harvested field (1\textsuperscript{st}, 4\textsuperscript{th}, 5\textsuperscript{th}, 7\textsuperscript{th} layer). Someone very tall is standing at the edge of the field, partially hidden behind the residue of the harvested crop (4\textsuperscript{th} layer). Considerable dust is generated in the headland area during the operation (4 \textsuperscript{th} layer).
    \item \label{itm:cult4} It is winter (5\textsuperscript{th} layer) as in \cref{fig:cultivate:sub4}. A pedestrian walks across a snow-covered field while an autonomous tractor works in parallel (4\textsuperscript{th}, 5\textsuperscript{th}, 7\textsuperscript{th} layer). There is a construction site at the edge of the field (3\textsuperscript{rd} layer).
\end{enumerate}

The defined parameters of the use case form the basis of the initial verification step. During this step, the preliminary Ag-ODD configuration 
is compared to the derived logical scenarios. This comparison identifies and refines any inconsistencies or boundary violations between the Ag-ODD and the logical scenarios. It is assumed that every modification to the Ag-ODD is based on a justified and documented reason. This ensures that all refinements are transparent and can be systematically traced throughout the iterative verification process \circled{4}. While these reasons are not stated explicitly here, they are implicitly assumed for each modification.

The initial Ag-ODD specifies in the \boxeded{scenery}{\sceneryColor} category, that all European fields are within the operational scope. However, the derived \cref{itm:cult2} reveals that areas with slopes of \SI{10}{\percent} or more are included. This is incorrect in our hypothetical assumption, because such conditions increase the difficulty of ensuring functional safety. Consequently, in \boxeded{LoD1}{\lodColor}, the attribute \emph{Fields in Europe} is refined to include only fields in Europe with slopes of up to \SI{10}{\percent}. The attribute remains permissive $\cup$, meaning all fields that meet this criterion are included. Furthermore, the initial \cref{itm:cult4} depict snow-covered ground, which should not be included in the Ag-ODD. This issue is addressed by refining the \emph{Uneven soil} attribute to explicitly exclude snow or frozen conditions. 

In the \boxeded{dynamic Objects}{\objectsColor} category, \emph{humans $\geq 2m$} within the operational scope are identified in \cref{itm:cult3} and must be excluded due to limitations of the imagined algorithms. This is because they could be classified as a tree. Additionally, the function is defined for a specific \emph{Tractor X} model and an implement width of $\leq$~\SI{10}{\meter}. 

All refined attributes, except for the \emph{Ego vehicle}, are assigned the type permissive $\cup$ because they are intended to include everything not explicitly described. These specifications are integrated into the refined Ag-ODD description, highlighted in blue cell background in \cref{tab:cultivation}.

Further verification is required to reach a clearly defined Ag-ODD \circled{2} that precisely describes the intended subject. Two additional logical scenarios \circled{3} are listed for comparison with the actual Ag-ODD.

\begin{enumerate}[label=\arabic*., ref=\arabic*, resume]
    \item \label{itm:cult5} As seen in \cref{fig:cultivate:sub5}, an autonomous tractor X cultivates a harvested field in an urban in Denmark under dense fog (1\textsuperscript{st}, 2\textsuperscript{nd}, 4\textsuperscript{th}, 5\textsuperscript{th}, 7\textsuperscript{th} layer). It uses a cultivator to perform cultivation work during twilight (4\textsuperscript{th}, 5\textsuperscript{th} layer).
    \item \label{itm:cult6} In \cref{fig:cultivate:sub6} Tractor X is preparing to cultivate a field next to a lake in Poland with a \SI{5}{\meter} wide implement (1\textsuperscript{st}, 2\textsuperscript{nd}, 4\textsuperscript{th}, 7\textsuperscript{th} layer). A man is sitting on an old well in the field (4\textsuperscript{th}, 5\textsuperscript{th} layer).
\end{enumerate}

Based on the logical \cref{itm:cult5} and logical scenario \cref{itm:cult6}, it is noted that collecting data from all possible environmental conditions in Europe would be too complex. Therefore, in this verification iteration highlighted by the cells with red cell background in \cref{tab:cultivation}, the scope will be narrowed down from \emph{Fields in Europe} and \emph{Slope $\leq$~\SI{10}{\percent}} to \emph{Fields in GER} and \emph{Slope $\leq$~\SI{10}{\percent}}. Additionally, the initial focus will be on implement Y by adjusting the cultivation implement attributes to \emph{Implement Y}. The \emph{Fields in GER} attribute is permissive because all German fields with a slope of $<$~\SI{10}{\percent} fall within the scope of the Ag-ODD. The \emph{Implement Y} attribute is assigned the restrictive type because it is described unambiguously, and other implements are not within the scope.
In addition, visibility is restricted in \cref{itm:cult5}, since safe operation cannot be guaranteed at visibility levels below \SI{50}{\meter}. Therefore, the \boxeded{environmental}{\environmentColor} conditions are also restricted.

In the context of this cultivation use and the verification process \circled{4} of \cref{fig:iterative_verification}, note that all newly introduced sub-attributes like \emph{Slope $\leq$~\SI{10}{\percent}} are interpreted as restrictive boundaries within the Ag-ODD. Therefore, all slopes of $\leq$~\SI{10}{\percent}  and all scenario boundaries pointing into the Ag-ODD are included. Values $>$~\SI{10}{\percent} are excluded from the Ag-ODD \circled{2} and scenario \circled{3} derivation. When an entire attribute is defined as restrictive, it indicates that its description is precise enough to establish a single, unambiguous boundary of inclusion or exclusion, with no gradation required.

Furthermore, the  \boxeded{process}{\processColor} attributes may be refined during this iterative procedure. If scenario analysis reveals that the defined process should be adjusted or extended, the corresponding attributes like start $x$ (SA$x$) condition $y$ (C$y$), and end z (EA$z$) must be updated, as well as the core process attribute itself.

This mechanism enables iterative verification of the Ag-ODD, ensuring a consistent and unique definition at all descriptive levels, including use cases, logical scenarios, and Ag-ODD attributes. As illustrated by the dashed lines in \cref{tab:cultivation}, the final Ag-ODD can be read from right to left. This procedure enables the Ag-ODD to be derived directly from the verified attributes, providing an unambiguous definition. The present example serves as a simplified illustration of this approach. In a practical application, the autonomous cultivation function would need to be validated within this uniquely defined Ag-ODD.

\subsubsection{Use Case 2: Wheat Harvesting}
\label{sec:ex_harvesting}
In general, \emph{harvesting} is often mentioned as the agricultural process with the most significant challenges when fully automating agricultural machinery. The complexity primarily arises from the possibility that people may be present in areas in front of the machine that are not directly visible. Consequently, the system's behavior depends heavily on its perception capabilities and subsequent sensor data processing. Achieving an optimal perception design, especially a reliable system behavior, requires a clear definition of the specific Ag-ODD \circled{2} wherein the autonomous function is intended to operate. Therefore, determining the Ag-ODD is a fundamental step in developing a safe and effective autonomous harvesting function, especially for use cases involving complex, safety-critical environments.

The wheat harvest use case is defined as follows:
\emph{Autonomous wheat harvesting is possible with continuous, 24/7 operation. Initially, trained personnel must harvest the headland to ensure a clear and detectable boundary between the field and its surroundings. Non-autonomous operations like crop transport and direct drilling can occur simultaneously within the same field. Human presence is permitted only inside agricultural machinery in the field. Responsibility for coordinating these machines lies with the operator supervising the autonomous system. The autonomous function is deactivated only in cases of heavy rain or dense fog. Other environmental and field parameters, such as slope, crop height, and soil type, are not expected to affect performance.}

\begin{figure*}[H]
    \centering
    \begin{subfigure}[b]{0.15\textwidth}
        \includegraphics[width=\textwidth]{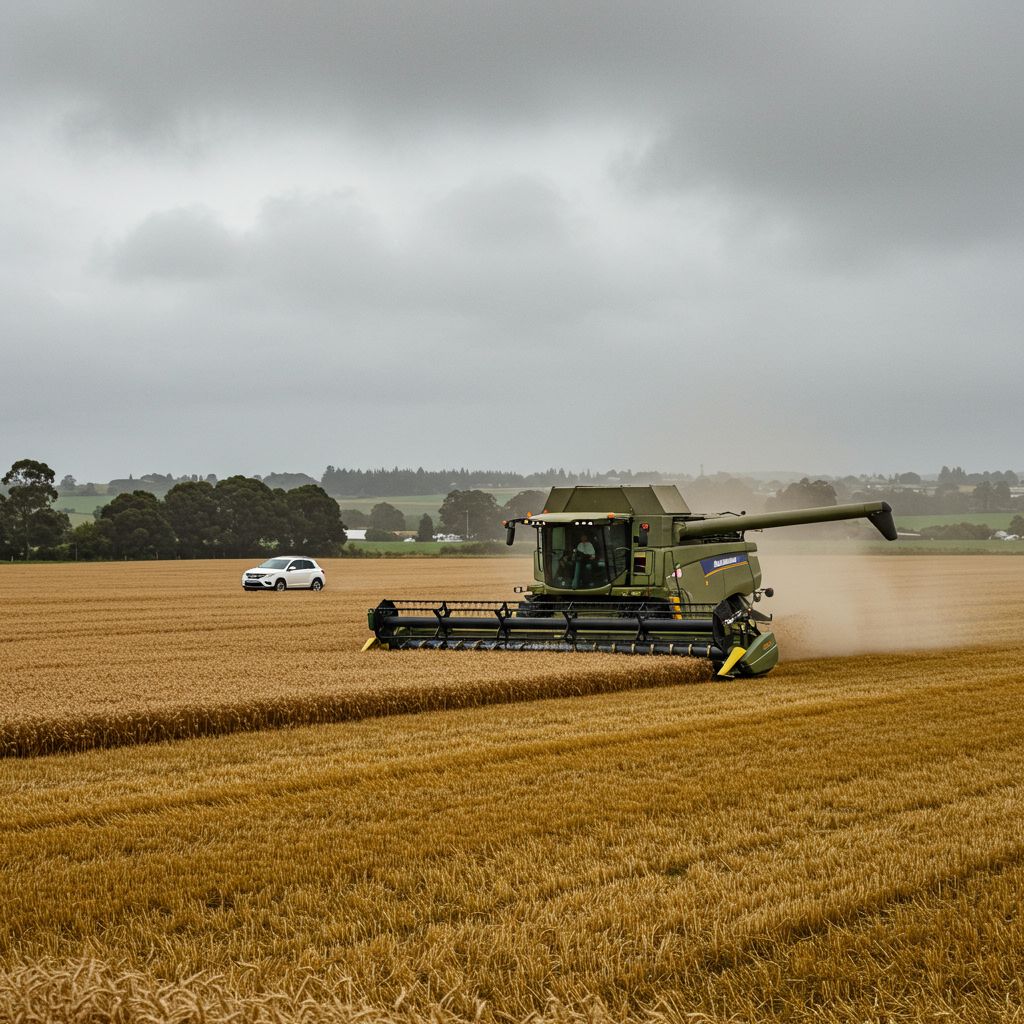}
        \caption{Scenario 1}
        \label{fig:wheat_harvest:sub1}
    \end{subfigure}\hfill
    \begin{subfigure}[b]{0.15\textwidth}
        \includegraphics[width=\textwidth]{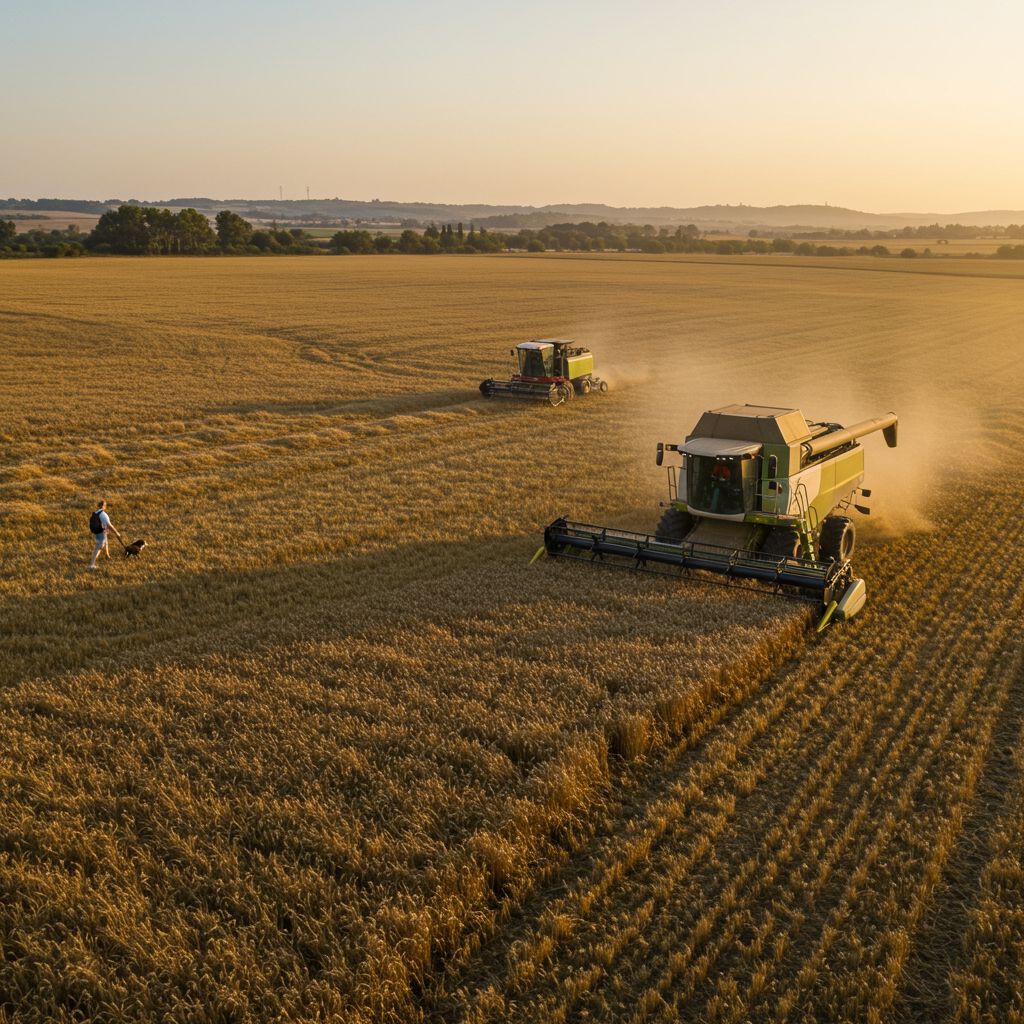}
        \caption{Scenario 2}
        \label{fig:wheat_harvest:sub2}
    \end{subfigure}\hfill
    \begin{subfigure}[b]{0.15\textwidth}
        \includegraphics[width=\textwidth]{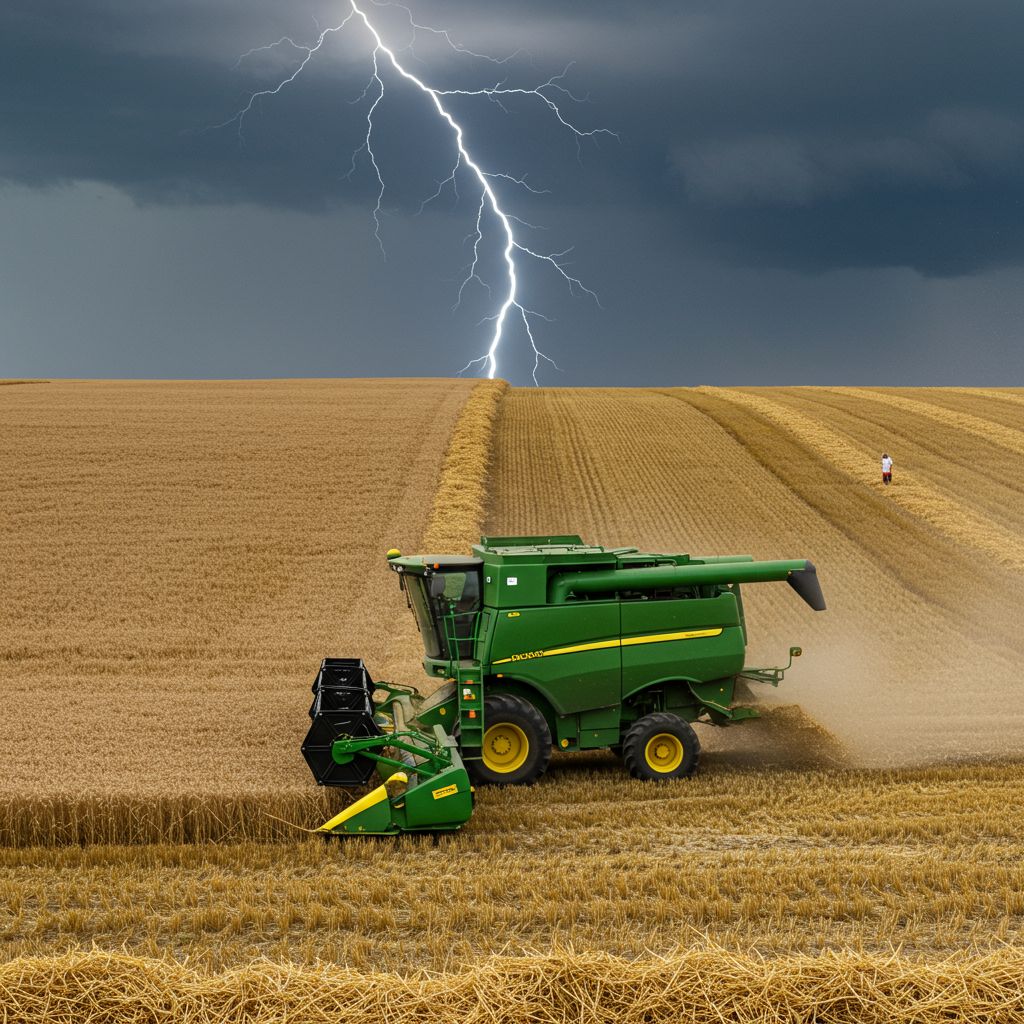}
        \caption{Scenario 3}
        \label{fig:wheat_harvest:sub3}
    \end{subfigure}\hfill
    \begin{subfigure}[b]{0.15\textwidth}
        \includegraphics[width=\textwidth]{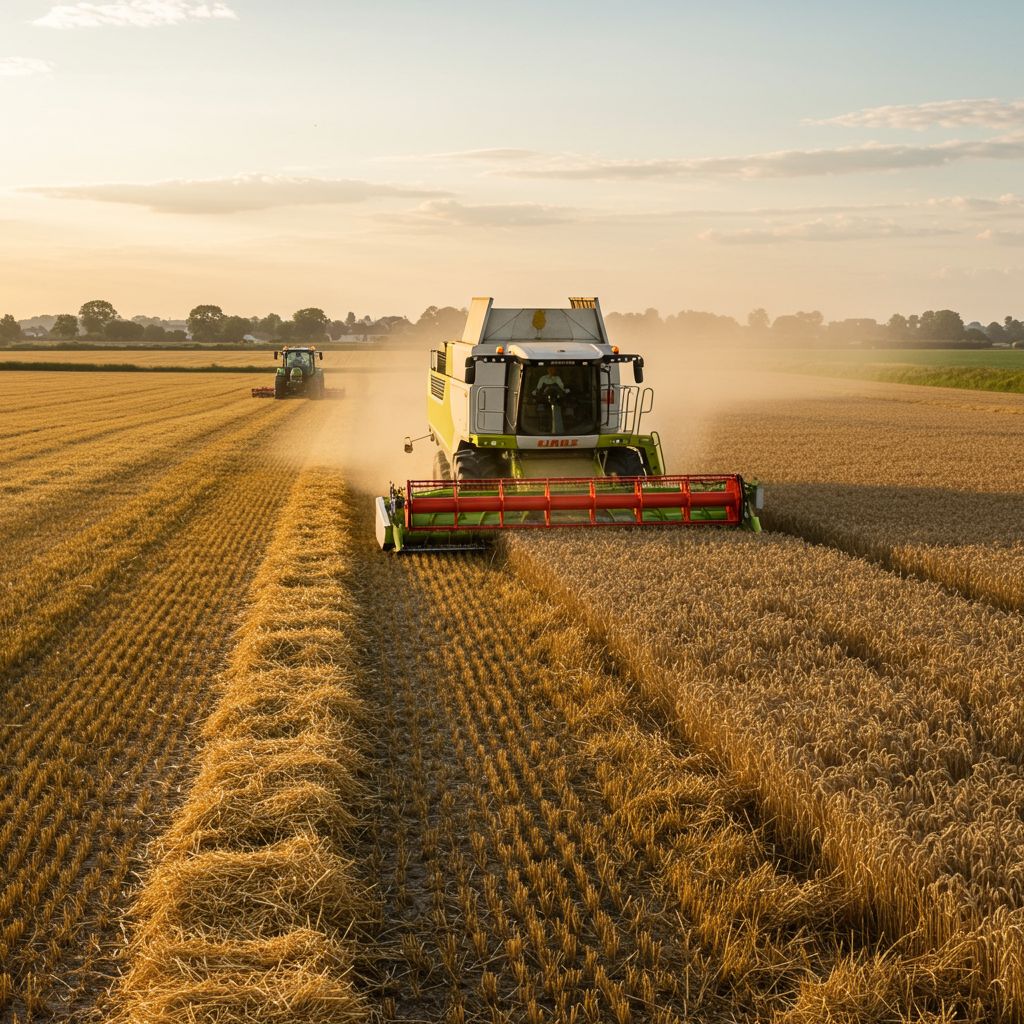}
        \caption{Scenario 4}
        \label{fig:wheat_harvest:sub4}
    \end{subfigure}\hfill
    \begin{subfigure}[b]{0.15\textwidth}
        \includegraphics[width=\textwidth]{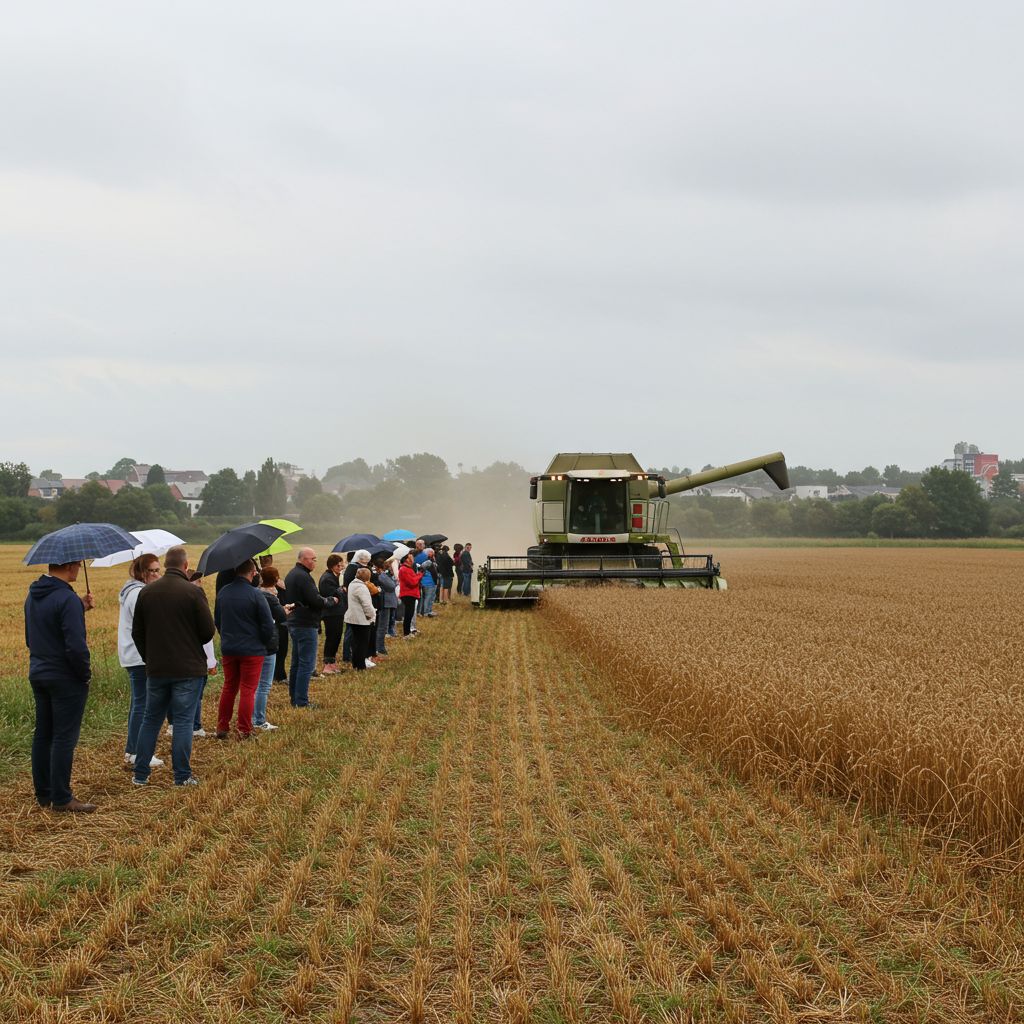}
        \caption{Scenario 5}
        \label{fig:wheat_harvest:sub5}
    \end{subfigure}\hfill
    \begin{subfigure}[b]{0.15\textwidth}
        \includegraphics[width=\textwidth]{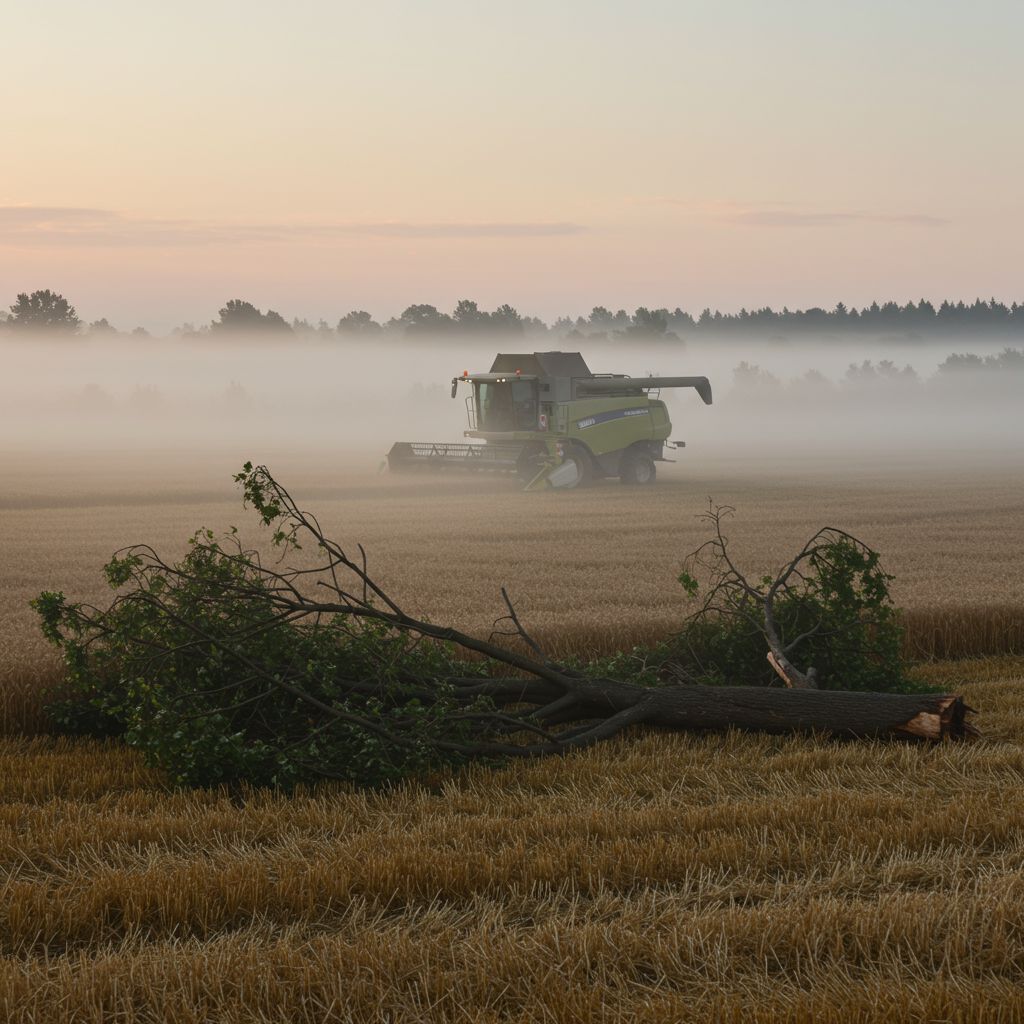}
        \caption{Scenario 6}
        \label{fig:wheat_harvest:sub6}
    \end{subfigure}

    \caption{Logical scenarios \protect\circled{3} derived using the 7-layer Model are presented here, in the example use case of \protect\emph{wheat harvesting}; these six different visualized scenarios are used during interactive verification \protect\circled{4}. The visualizations (\protect\subref{fig:wheat_harvest:sub1}) to (\protect\subref{fig:wheat_harvest:sub6}) are AI-generated images.}
    \label{fig:wheat_harvest_images}
\end{figure*}

The initial structuring of this use case within the Ag-ODD framework is shown in \cref{tab:wheat_harvest} as cells with white background color. Several particularities are evident in this example use case \circled{1}. First, the Ag-ODD \circled{2} description incorporates multiple \boxeded{processes}{\processColor} that interact with individual attributes. These processes include not only the autonomous harvesting process itself, but also parallel non-autonomous operations, such as direct drilling. Note that the direct drilling process can directly alter the environment, whereas crop transportation cannot. Additionally, it is evident that there are already established \boxeded{LoD}{\lodColor} that allow for a more precise description of the Ag-ODD. These aspects will be verified and specified in more detail, or narrowed down by comparing them with derived logical scenarios \circled{3}. To illustrate the verification process \circled{4} for the wheat harvest use case, six representative logical scenarios \circled{3} are defined as shown in \cref{fig:wheat_harvest_images}. These scenarios are designed to test the boundaries of the initially defined Ag-ODD and to identify conditions that either fall outside the valid operational domain or require refinement of its parameters. As in \cref{sec:ex_cultivation}, the layer numbers for creating scenarios are written in brackets after the lingual descriptions. 

\begin{table*}[tbh]
\centering
\renewcommand{\arraystretch}{1.2}
\caption{The Ag-ODD for the Wheat Harvesting use case is shown below. The \protect\boxeded{process}{\processColor} category generates condition-dependent variables (CDV) with start attribute SA1, condition C1, and end attribute EA1. The descriptions of the \protect\boxeded{scenery}{\sceneryColor}, the \protect\boxeded{environment}{\environmentColor} and the \protect\boxeded{dynamic objects}{\objectsColor} categories are represented with different \protect\boxeded{levels of detail (LoD)}{\lodColor}. The first iteration is highlighted with white and the second with blue cell background. The \protect\boxeded{permissive/restrictive attribute properties}{\propColor}, shows whether an attribute is permissive ($\cup$) or restrictive ($\cap$) is in the Type column. The Ag-ODD is by definition, entirely restrictive---anything not listed is excluded. The current or final Ag-ODD is represented from the right by the dashed line.}

\scalebox{0.84}{
\begin{NiceTabular}
{>{\raggedright\arraybackslash}p{0.25\textwidth} 
>{\centering\arraybackslash}p{0.1\textwidth} 
>{\centering\arraybackslash}p{0.1\textwidth} 
>{\raggedright\arraybackslash}p{0.25\textwidth} 
>{\centering\arraybackslash}p{0.1\textwidth} 
>{\centering\arraybackslash}p{0.1\textwidth} 
}
\boxeded{LoD0}{\lodColor}                                    & \textbf{}          & \boxeded{Type}{\propColor}         & \multicolumn{1}{||l}{\boxeded{LoD1}{\lodColor} }                                                  &     & \boxeded{Type}{\propColor}               \\ \hline
\boxeded{scenery}{\sceneryColor}                                   & \textbf{}          & $\cap$           & \multicolumn{1}{||l}{}                                                               &     &                              \\ 
                                                   &                    &                       & \multicolumn{1}{||l}{Crop height (\SI{40}{\centi\meter} -- \SI{60}{\centi\meter})}                                    & SA1    & $\cap$                    \\ 
                                                   &                    &                       & \multicolumn{1}{||l}{Crop height harvested (\SI{5}{\centi\meter} -- \SI{25}{\centi\meter})}                           & EA1/SA2& $\cap$                     \\ 
                                                   &                    &                       & \multicolumn{1}{||l}{\cellcolor[HTML]{FFCCC9}Grain moisture content $\leq y$ \%}                           & \cellcolor[HTML]{FFCCC9} & \cellcolor[HTML]{FFCCC9}$\cup$                     \\
                                                   &                    &                       & \multicolumn{1}{||l}{\cellcolor[HTML]{FFCCC9}Biomass of the crop in $\mathrm{kg}$}                           & \cellcolor[HTML]{FFCCC9}& \cellcolor[HTML]{FFCCC9}$\cup$                     \\
                                                   &                    &                       & \multicolumn{1}{||l}{\cellcolor[HTML]{FFCCC9}Yield \unit{\tonne\per\hectare}}                           & \cellcolor[HTML]{FFCCC9}& \cellcolor[HTML]{FFCCC9}$\cup$                     \\
\multirow{-6}{*}{Crop}                             & \multirow{-6}{*}{EA2} & \multirow{-6}{*}{$\cap$ } & \multicolumn{1}{||l}{\cellcolor[HTML]{CBCEFB}Crop type winter wheat}                    & \cellcolor[HTML]{CBCEFB} & \cellcolor[HTML]{CBCEFB}$\cap$ \\
Soil                                               &                    & $\cup$                    & \multicolumn{1}{||l}{}                                                                &  &                                \\ 
                                                   &                    &                       & \multicolumn{1}{||l}{\cellcolor[HTML]{CBCEFB}Slope $\leq x$}                          & \cellcolor[HTML]{CBCEFB} & \cellcolor[HTML]{CBCEFB}$\cup$        \\ 
\multirow{-2}{*}{Fields}                           & \multirow{-2}{*}{} & \multirow{-2}{*}{$\cup$ } & \multicolumn{1}{||l}{\cellcolor[HTML]{CBCEFB}Fields in Portugal}                      & \cellcolor[HTML]{CBCEFB} & \cellcolor[HTML]{CBCEFB}$\cup$        \\
Surroundings                                       &                    & $\cup$                    & \multicolumn{1}{||l}{}                                                               &     &                                \\ \hline
\boxeded{environment}{\environmentColor}                                &                    & \textbf{$\cap$ }          & \multicolumn{1}{||l}{}                                                               &     &                                \\ 
                                                     &                    &                       & \multicolumn{1}{||l}{\cellcolor[HTML]{FFCCC9}Relative humidity $\leq z$ \%}                                                   & \cellcolor[HTML]{FFCCC9}    & \cellcolor[HTML]{FFCCC9}$\cup$   \\ 
                                                   &                    &                       & \multicolumn{1}{||l}{No dense fog (visibility $\leq$~\SI{25}{\meter})}                                                   &     & $\cup$   \\  
                                                   &                    &                       & \multicolumn{1}{||l}{No heavy rain (visibility $\leq$~\SI{25}{\meter})}                                                  &     & $\cup$   \\  
\multirow{-4}{*}{All wheather conditions}          & \multirow{-4}{*}{} & \multirow{-4}{*}{$\cup$ } & \multicolumn{1}{||l}{\cellcolor[HTML]{CBCEFB}No dense dust (visibility $\leq$~\SI{50}{\meter})}       & \cellcolor[HTML]{CBCEFB}  & \cellcolor[HTML]{CBCEFB}$\cup$  \\ \hline
\boxeded{dynamic objects}{\objectsColor}                                   &                    & $\cap$           & \multicolumn{1}{||l}{}                                                               &     &      \\ 
\cellcolor[HTML]{CBCEFB}                           &\cellcolor[HTML]{CBCEFB}& \cellcolor[HTML]{CBCEFB} & \multicolumn{1}{||l}{\cellcolor[HTML]{CBCEFB}Humans}                & \cellcolor[HTML]{CBCEFB} & \cellcolor[HTML]{CBCEFB}$\cup$   \\ 
\multirow{-2}{*}{\cellcolor[HTML]{CBCEFB}Humans}   &\multirow{-2}{*}{\cellcolor[HTML]{CBCEFB}} & \multirow{-2}{*}{\cellcolor[HTML]{CBCEFB}$\cup$ } & \multicolumn{1}{||l}{\cellcolor[HTML]{CBCEFB}Humans with objects}   &  \cellcolor[HTML]{CBCEFB} & \cellcolor[HTML]{CBCEFB}$\cup$  \\ 
Animals                                            &                    & $\cup$                    & \multicolumn{1}{||l}{}                                                               &     &                                \\ 
                                                   &                    &                       & \multicolumn{1}{||l}{Ego-vehicle}                                                    & C1  & $\cup$                             \\ 
\multirow{-2}{*}{Vehicle}                          &\multirow{-2}{*}{}  & \multirow{-2}{*}{$\cup$ } & \multicolumn{1}{||l}{\cellcolor[HTML]{CBCEFB}Other vehicle}                         &  \cellcolor[HTML]{CBCEFB} & \cellcolor[HTML]{CBCEFB}$\cup$     \\ 
Agricultural machinery                             &                    & $\cup$                    & \multicolumn{1}{||l}{\cellcolor[HTML]{CBCEFB}Distance to ego-vehicle $\geq D$}      &  \cellcolor[HTML]{CBCEFB} & \cellcolor[HTML]{CBCEFB}$\cup$     \\ 
Agricultural implements                            &C2                  & $\cup$                    & \multicolumn{1}{||l}{\cellcolor[HTML]{CBCEFB}Distance to ego-vehicle $\geq D$}      &  \cellcolor[HTML]{CBCEFB} C2.1& \cellcolor[HTML]{CBCEFB}$\cup$ \\  \hline
& & & & & \\[-2ex]  
\boxeded{process}{\processColor}                                  & \multicolumn{2}{c}{\textbf{Start}}                                & \multicolumn{2}{c}{\textbf{Condition}}           & \textbf{End}         \\ \hline
24/7 autonomous harvesting                         & \multicolumn{2}{c}{SA1}                                           & \multicolumn{2}{c}{Interaction with C1}          & EA1 \\
24/7 cultivation                                   & \multicolumn{2}{c}{SA2}                                           & \multicolumn{2}{c}{Interaction with C2 C2.1}     & EA2 \\ 
24/7 crop transport                                & \multicolumn{2}{c}{---}                                           & \multicolumn{2}{c}{---}     & --- \\         

\CodeAfter
\tikz \draw [line width=2pt, black, dash dot dot] (3-|last) |- (3-|4) |- (9-|4) |- (9-|1) |- (10-|1) |- (10-|4) |- (12-|4) |- (12-|1) |- (13-|1) |- (13-|4) |- (21-|4) |- (21-|1) |- (22-|1) |- (22-|4) |- (26-|4) |- (26-|last) |- (22-|last) |- ($(22-|4)+(0.2,0)$) |- ($(21-|4)+(0.2,0)$) |- (21-|last) |- ($(10-|4)+(0.2,0)$) |- ($(9-|4)+(0.2,0)$) |- (9-|last) -- cycle;
\end{NiceTabular}
}

\label{tab:wheat_harvest}
\end{table*}

\begin{enumerate}[label=\arabic*., ref=\arabic*]
    \item \label{itm:wheat1} An autonomous combine harvester operates in a wheat field in New Zealand as show as in \cref{fig:wheat_harvest:sub1} (1\textsuperscript{st}, 4\textsuperscript{th}, 5\textsuperscript{th}, 7\textsuperscript{th} layer). Light rain begins to fall, and apart from a car parked at the edge of the field, there are no other objects or people present (2\textsuperscript{nd}, 4\textsuperscript{th}, 5\textsuperscript{th} layer). Relative humidity is high (5\textsuperscript{th} layer).
    \item \label{itm:wheat2} An autonomous combine harvester is harvesting wheat in Portugal (1\textsuperscript{st}, 4\textsuperscript{th}, 5\textsuperscript{th}, 7\textsuperscript{th} layer). As visualized in \cref{fig:wheat_harvest:sub2}, in parallel, a tractor cultivates within the same field (4\textsuperscript{th}, 7\textsuperscript{th} layer). A pedestrian with a dog crosses the opposite side of the field (4\textsuperscript{th} layer). The low sun is partially obscured by surrounding trees and may cause glare (2\textsuperscript{nd}, 5\textsuperscript{th} layer).
    \item \label{itm:wheat3} A red combine harvester sits at the edge of a steeply sloped field of lodged grain (2\textsuperscript{nd}, 4\textsuperscript{th}, 7\textsuperscript{th} layer). The upper body of a person becomes visible behind the already harvested headland (2\textsuperscript{nd}, 4\textsuperscript{th} layer). As depicted in \cref{fig:wheat_harvest:sub3}, thunderstorm activity is expected soon (5\textsuperscript{th} layer).
    \item \label{itm:wheat4} A green combine harvester operates autonomously along its harvesting lane (2\textsuperscript{nd}, 4\textsuperscript{th}, 7\textsuperscript{th} layer) shown in \cref{fig:wheat_harvest:sub4}. On its left, the field has already been harvested, and a tractor approaches in the opposite direction for cultivation (1\textsuperscript{at}, 4\textsuperscript{th}, 7\textsuperscript{th} layer). The tractor generates a dense dust cloud behind it, while a road borders the field edge (2\textsuperscript{nd}, 5\textsuperscript{th} layer).
    \item \label{itm:wheat5} An autonomous combine harvester, shown in \cref{fig:wheat_harvest:sub5}, is harvesting a field on the outskirts of the city (1\textsuperscript{st}, 2\textsuperscript{en}, 4\textsuperscript{th}, 7\textsuperscript{th}layer). Spectators gather along the edge of the field to observe the process (4\textsuperscript{th}layer). Several people are holding umbrellas due to light rain (4\textsuperscript{th}, 5\textsuperscript{th} layer).
    \item \label{itm:wheat6} As illustrated in \cref{fig:wheat_harvest:sub6}, a combine harvester is harvesting a wheat field (7\textsuperscript{th} layer). It is following a pre-planned mission with a route and tasks (6\textsuperscript{th} layer). It is early in the day and foggy, so the grain moisture content is too high to harvest (1\textsuperscript{st}, 5\textsuperscript{th} layer).
    There is a large fallen tree at the edge of the field (3\textsuperscript{rd} layer).
\end{enumerate}

As before, it is assumed that any modification to the Ag-ODD is based on a justified and documented reason. This ensures that all refinements are transparent and can be systematically traced throughout the iterative verification process \circled{4}. While these reasons are not explicitly stated here, they are implicitly assumed for each modification. 

The six scenarios suggest that the initial Ag-ODD is not yet fully developed. The first two scenarios demonstrate that the current use case includes agricultural fields worldwide. Thus, the scope is too broad for practical implementation, as there is no worldwide machine certification. To ensure a certifiable, manageable and well-defined Ag-ODD, therefore, the geographical extent must be narrowed to a specific region, such as \emph{Fields in Portugal}.

In the second iteration, as indicated by the blueish cells in the \cref{tab:wheat_harvest}, the \emph{Fields} attribute incorporates two additional attributes in the second level \boxeded{LoD1}{\lodColor}. First, the \emph{Slope $\leq$~x} is defined with the threshold x; second, only \emph{Fields in Portugal} are considered. Both parameters are permissive because they only constrain the external boundaries of the attribute, not its internal pointing boundaries. They allow for further specification. Additionally, only one type of wheat should be processed. Subsequently, the restrictive attribute \emph{Crop type winter wheat} in \boxeded{LoD1}{\lodColor} under the attribute \emph{Crop} is added to ensure its unambiguity.

It is also revealed by the first scenario, cf. \cref{itm:wheat1} that not only agricultural machinery, but also other types of vehicles, may be present within the use case. Therefore, if such occurrences are to be included, a new \boxeded{dynamic object}{\objectsColor} attribute named \emph{Other vehicle} is introduced in the second iteration. Alternatively, these vehicles could be excluded entirely, requiring the introduction of a new attribute such as \emph{No other vehicle}.

The presence of humans in the field, as shown in \cref{itm:wheat2} and \cref{itm:wheat5}, in various forms, cannot be effectively prevented under normal operating conditions. Accordingly, \emph{Humans} are incorporated as an additional attribute. At \boxeded{LoD1}{\lodColor}, this attribute is refined to distinguish between people carrying objects, e.g., umbrellas, and those who are not. All three of the newly introduced attributes are permissive ($\cup$), as no further limitation of the humans is mentioned in the scenarios \circled{3}.

Dense dust is not a desirable condition,  as it behaves  similar to fog. Consequently, \emph{No dense dust} is added as a new attribute at \boxeded{LoD1}{\lodColor}. Finally, the new \emph{Distance to ego-vehicle $\geq D$} attribute adjusts the minimum distance between other machines working in the field and the autonomous system. 

This change eliminates the need predictive path modeling of other vehicles, which may facilitate subsequent simulation effort in terms of resources and computing capacities.
However, as noted in the previous use case, performing the iterative verification until \cref{tab:wheat_harvest}  is a well-defined Ag-ODD is beyond the scope of this work.

Up to this point, the example has served as another illustration of the Ag-ODD definition for autonomous driving. However, since the process is the primary focus in the case of the combine harvester, i.e., driving is process-driven, the process should be considered. Since a new function, namely grain harvesting process, is now being considered, there is no need to create new scenarios; instead, the existing ones can be used. The newly 
acquired attributes are nevertheless entered as red cells in \cref{tab:wheat_harvest} as a further iteration.
In all given scenarios, in order to operate the combine harvester as efficiently as possible at its performance limit, the forward speed during harvesting must be adjusted according to the biomass of the crop. In addition, a certain yield should be achieved.
The high humidity caused by a change in weather conditions prevents harvesting from continuing as shown in \cref{itm:wheat1}.
The situation is similar in \cref{itm:wheat6}, where excessive grain moisture content prevents harvesting from beginning.
It is therefore necessary to set limits for grain moisture content and relative humidity.

Based on this principle, one can derive that a single Ag-ODD may encompass multiple sub-Ag-ODDs. Each sub-Ag-ODD represents a distinct machine process or functional domain. Examples include autonomous driving and implement control. The structured use of LoDs allows for flexible navigation within the description, enabling a zoom-in to include more specific attributes for a given function or a zoom-out to consider a broader operational context. This approach ensures consistent, unambiguous representation of multiple functions within a unified Ag-ODD structure.

\subsection{General Considerations and Fundamental Recommendations}
\label{sec:general_considerations}
By two example use cases, \cref{sec:ex_cultivation} and \cref{sec:ex_harvesting}, the method of derivation and iterative verification \circled{4} of an Ag-ODD \circled{2} is exemplified. Both are used to demonstrate the general process of the Ag-ODD Framework and are not intended to derive a complete, usable Ag-ODD for one of these use cases; they are highly simplified. Scaling that towards an Ag-ODD for production machinery suggests that the process will have high complexity. Still, due to its structure, it allows traceability and has the potential to generate constructive and nearly gapless Ag-ODD.

In general, new attributes may be introduced at any \boxeded{LoD}{\lodColor} during each verification iteration. During these iterations, it is recommended to add new attributes rather than merge them. For instance, in the initial structured description of the Ag-ODD, \emph{Vehicle} and \emph{Agricultural machinery} may be listed as separate attributes rather than \emph{Agricultural machinery} being listed as a \boxeded{LoD}{\lodColor} of the \emph{Vehicle} attribute. This approach preserves the evolutionary iteration steps and ensures the traceability of the Ag-ODD's development. The Ag-ODD can be considered verified once no further logical scenarios can be identified that would necessitate modifications to it.

After the iterative verification process the Ag-ODD in tabular form, as exemplarily performed in \cref{tab:cultivation} and \cref{tab:wheat_harvest} , the resulting analysis should be read from right to left.

After verification of the Ag-ODD in tabular form,  as exemplarily performed in \cref{tab:cultivation} and \cref{tab:wheat_harvest} and suggested by other application of functional safety \citep{Jacobs2013,Komesker2025} , the resulting analysis should be read from right to left. Major boundaries is indicated by the dotted line. The different \boxeded{LoD}{\lodColor} (LoD1-LoD$X$) progressively constrain the corresponding LoD0 attribute.

Attributes can be associated with specific processes. When processed in that manner, all of that attribute's more detailed LoD \boxeded{permissive/restrictive attribute properties}{\propColor}  also apply to the process itself. In other words, a process is only considered part of the Ag-ODD if it satisfies all higher-level attribute constraints.

\subsection{Evaluation and Validation of the Function within the Agricultural Operational Design Domain}
\label{sec:val_ex_fun}
Validating the function is the most critical and challenging phase in the development of autonomous driving systems for the agricultural domain. The framework presented in this work for deriving a verified Ag-ODD \circled{2} serves as the foundation for the subsequent validation stage based on a single or multiple use cases \circled{1}. Effective validation requires the generation of concrete scenarios with sufficient variance, derived from the logical scenarios that comprehensively represent the Ag-ODD following the iterative verification process.

For example, if the Ag-ODD includes the presence of people in a logical scenario \circled{3}, the corresponding concrete scenarios must also include a realistic variance of human representation to capture the variability observed under real-world conditions. This includes variations in skin tone, body size, clothing, and behavior. Ensuring such realism is crucial for producing meaningful, robust validation results.

Furthermore, as proposed by \citet{Happich2025}, the validation process should align with established and emerging standards. The standard ensures that the validation process involves identifying and applying relevant parameters and test procedures alternating from sensor-level, subsystem, and full-system tests derived from the defined Ag-ODD. Linking these tests systematically to the verified Ag-ODD and its associated functional requirements enables evaluation of the autonomous function's reliability and safety within its intended agricultural operational design domain.

Logically, validation strategies are generally not limited to testing a functional behavior within normal working conditions. It is essential to test the limits of the Ag-ODD in both directions: the logical scenarios \circled{3} and against itself \circled{2} . This means the logical scenarios should intentionally extend and stress beyond the defined Ag-ODD to examine the function's behavior under boundary and out-of-scope conditions. If the automated functions fail to operate correctly within the defined Ag-ODD, then the underlying assumptions must be revisited, including the framing limitations I) functional requirements, defined II) system capabilities, and III) the HARA. The iterative derivation and verification process ensures that the Ag-ODD and the autonomous function evolve consistently toward a safe and reliable design matching the designated use cases.

\section{Conclusion}
This work presents an Ag-ODD Framework as a contribution to the safe and structured development of autonomous agricultural machinery. Existing ODD structures were reviewed in \cref{sec:SOTA} for their applicability to agriculture, revealing in \cref{sec:synopsis and synthesis} that current approaches lack adequate mechanisms to represent dynamic agricultural processes or to describe an Ag-ODD unambiguously and comprehensively for use in such contexts.

The proposed framework is based on the ASAM OpenODD structure and builds on it by introducing an operational context category, which incorporates both process and scenery categories. It also integrates the LoD concept from CityGML and distinguishes between permissive/restrictive attribute properties, providing a consistent and transparent foundation for defining Ag-ODDs.

An iterative verification process has been established to demonstrate how the PEGASUS method for generating logical scenarios can be adapted for agricultural applications. Full coverage and verification can be achieved through iterative comparison between the Ag-ODD and its logical scenarios. Two illustrative use cases demonstrate the applicability and benefits of the framework in practice.

The Ag-ODD Framework can be used to develop unambiguous, verifiable Ag-ODDs for autonomous and semi-autonomous functions in any context. Although it was designed for off-road and agricultural environments, its reliance on ASAM OpenODD and PEGASUS ensures that it is also compatible with on-road contexts. Its modular structure enables it to be easily integrated into existing development workflows and simulation interfaces.

This work is regarded as foundational for the future development of autonomous agricultural functions and is an integral part of the overall development process. To ensure the consistency, verifiability, and alignment of the Ag-ODD and its function with current and emerging standards, each use case and function can undergo the described steps independently. Given that the majority of standards place significant emphasis on the necessity of explicitly delineating the ODD as a prerequisite for ensuring safety assurance, this framework is imperative for fulfilling those requirements.

\bibliography{ref.bib}

\end{document}